\documentclass{article} 
\usepackage[final]{colm2026_conference}

\usepackage{microtype}
\usepackage{hyperref}
\usepackage{url}
\usepackage{booktabs}
\usepackage{wrapfig}
\usepackage{amsmath}
\usepackage{amssymb}
\usepackage{mathtools}
\usepackage{amsthm}
\usepackage{makecell}
\usepackage{graphicx}
\usepackage{subcaption}
\usepackage{xcolor}
\usepackage{multirow}
\usepackage{multicol}
\usepackage{pifont}
\usepackage[misc]{ifsym}

\usepackage{color}
\usepackage[table]{xcolor}
\definecolor{mygreen}{RGB}{0,120,90}

\usepackage[capitalize,noabbrev]{cleveref}

\definecolor{darkblue}{rgb}{0, 0, 0.5}
\hypersetup{colorlinks=true, citecolor=mid, linkcolor=primary, urlcolor=mid}

\renewcommand{\cite}[1]{\citep{#1}}

\usepackage{xspace}

\usepackage{graphicx}
\usepackage{xcolor}
\usepackage{caption}
\usepackage{tcolorbox}
\tcbuselibrary{skins,breakable,listings}
\usepackage{listings}
\usepackage{inconsolata} 
\usepackage[T1]{fontenc}       
\usepackage{inconsolata}       

\usepackage{multirow}
\usepackage{booktabs}
\usepackage{makecell}
\usepackage{float}
\usepackage{algorithm}
\usepackage{algpseudocode}
\usepackage{fontawesome5}
\definecolor{thinkbg}{HTML}{E7F5FF}  
\definecolor{darkblue}{HTML}{bac8ff}
\definecolor{darkyellow}{HTML}{fff9db}
\newtcolorbox[auto counter,number within=section]{prompttemplate}[2][]{
  colback=gray!10,
  colframe=black!70,
  boxrule=0.6pt,
  arc=4pt,
  left=6pt,right=6pt,top=6pt,bottom=6pt,
  title={Template~\thetcbcounter: #2},
  #1
}

\newtcolorbox{myframe}{
  colback=white,
  colframe=black,
  arc=6pt,              
  boxrule=0.6pt,        
  left=3pt,
  right=3pt,
  top=3pt,
  bottom=3pt,
  enhanced
}
\newtcolorbox{promptbox}[1][]{
  enhanced,
  breakable,
  colback=white,          
  colframe=black!25,
  colbacktitle=black!8,   
  coltitle=black,
  fonttitle=\bfseries,
  boxrule=0.5pt,
  title={#1}
}

\newtcolorbox{thinkcodebox}[1][]{
  enhanced,
  breakable,
  colback=thinkbg,            
  colframe=black!25,
  colbacktitle=black!8,   
  boxrule=0.5pt,
  title={#1},
  fonttitle=\bfseries,
  coltitle=black
}

\lstdefinestyle{modelcode}{
  basicstyle=\ttfamily\small,
  columns=fullflexible,
  breaklines=true,
  backgroundcolor=\color{black!5},  
  frame=none
}

\newtcolorbox{sandboxbox}[1][]{
  enhanced,
  breakable,
  colback=darkyellow,
  colframe=black!25,
  colbacktitle=black!8,   
  boxrule=0.5pt,
  title={#1},
  fonttitle=\bfseries,
  coltitle=black
}

\newtcolorbox{answerbox}[1][]{
  enhanced,
  breakable,
  colback=darkblue,            
  colframe=black!25,
  colbacktitle=black!8,   
  boxrule=0.5pt,
  title={#1},
  fonttitle=\bfseries,
  coltitle=black
}

\title{\emph{FaithEyes}: Towards Faithful Tool Use via Multi-Agent Process-Image Self-Verification}

\author{
\mbox{Haoqing Wang$^{1}$, Xingrun Xing$^{1}$, Ziheng Li$^{2}$, Jianyuan Guo$^{3}$, Yehui Tang$^{1{~\textrm{\Letter}}}$}\\
$^1$ Samsung Research, Beijing, China \quad $^2$ Peking University \quad $^3$ CityUHK \\
\texttt{\{haoqing.wang, yehui.tang\}}@samsung.com \\
$^\textrm{\Letter}$~Corresponding Author
}

\begin{document}

\maketitle
\begin{abstract}
Agentic vision-language models (VLMs), which interleave textual reasoning with explicit tool calls such as cropping and code-based image manipulation, have emerged as a compelling paradigm for reliable and interpretable multi-modal reasoning. However, recent studies have revealed that such models often use tools unfaithfully. Many process images are irrelevant to the question (e.g., the crops miss the queried target), yet the tool call still receives full credit and the model still answers correctly. Such decorative or misaligned tool calls waste computation and reveal that the model does not faithfully use the evidence it retrieves. This may stem from two limitations of prevailing methods: the tool reward fails to distinguish useful from useless calls, and tool feedback carries no signal of usefulness. To this end, we introduce \emph{\textbf{FaithEyes}}, a multi-agent self-judging framework. Concretely, we use a VLM to judge whether each process image helps answer the question. The judgement is injected into the reasoning context as part of the tool observation to help subsequent reasoning, and meanwhile is used to scale the tool reward by the helpful-tool ratio to suppress reward hacking. To keep judgement available at evaluation, we further design a multi-agent framework where the model itself serves as a subagent to judge the tool calls from the main agent, eliminating any dependence on external models at inference. Training via a two-stage SFT + RL pipeline on adapted open-source data, FaithEyes attains competitive or superior accuracy across visual perception and reasoning benchmarks, while substantially improving tool faithfulness and reducing inference cost. The homepage is at \url{https://github.com/Mosi-AI/FaithEyes}. 
\end{abstract}

\section{Introduction}
Recent advances in agentic VLMs have demonstrated remarkable potential for multi-modal reasoning and perception. By integrating tool invocation into the reasoning process, these models can actively manipulate visual inputs (such as cropping, zooming, and rotating images) and retrieve supplementary information through code execution or web search, thereby achieving reliable and interpretable problem solving~\citep{zheng2025deepeyes,zhang2025thyme,lai2025mini}. Notably, compact agentic VLMs can surpass significantly larger models on challenging benchmarks~\citep{zheng2025deepeyes}. These results underscore the research value of enhancing VLMs with agentic tool-use capabilities.

Despite these promising results, a growing body of recent works has revealed that agentic VLMs often use tools unfaithfully~\citep{liu2025faithful,hou2026codev,yang2026position}. A prominent symptom is that the tool operates on the wrong evidence. Even when the final answer is correct, only about half of the samples actually contain at least one tool call cropping the region the question asks about~\citep{hou2026codev}, while the rest return decorative or misaligned process images. Such off-target calls co-occurring with correct answers may stem from that the model can shortcut through prior knowledge or even random guessing. Consequently, the tool invocation loses its intended purpose, degenerating into reward hacking where the model calls tools without meaningfully engaging with their outputs. This wastes computational resources on unnecessary process images for simple questions while failing to develop robust visual reasoning for complex ones.

\begin{figure}[t]
    \centering
    \includegraphics[width=\linewidth]{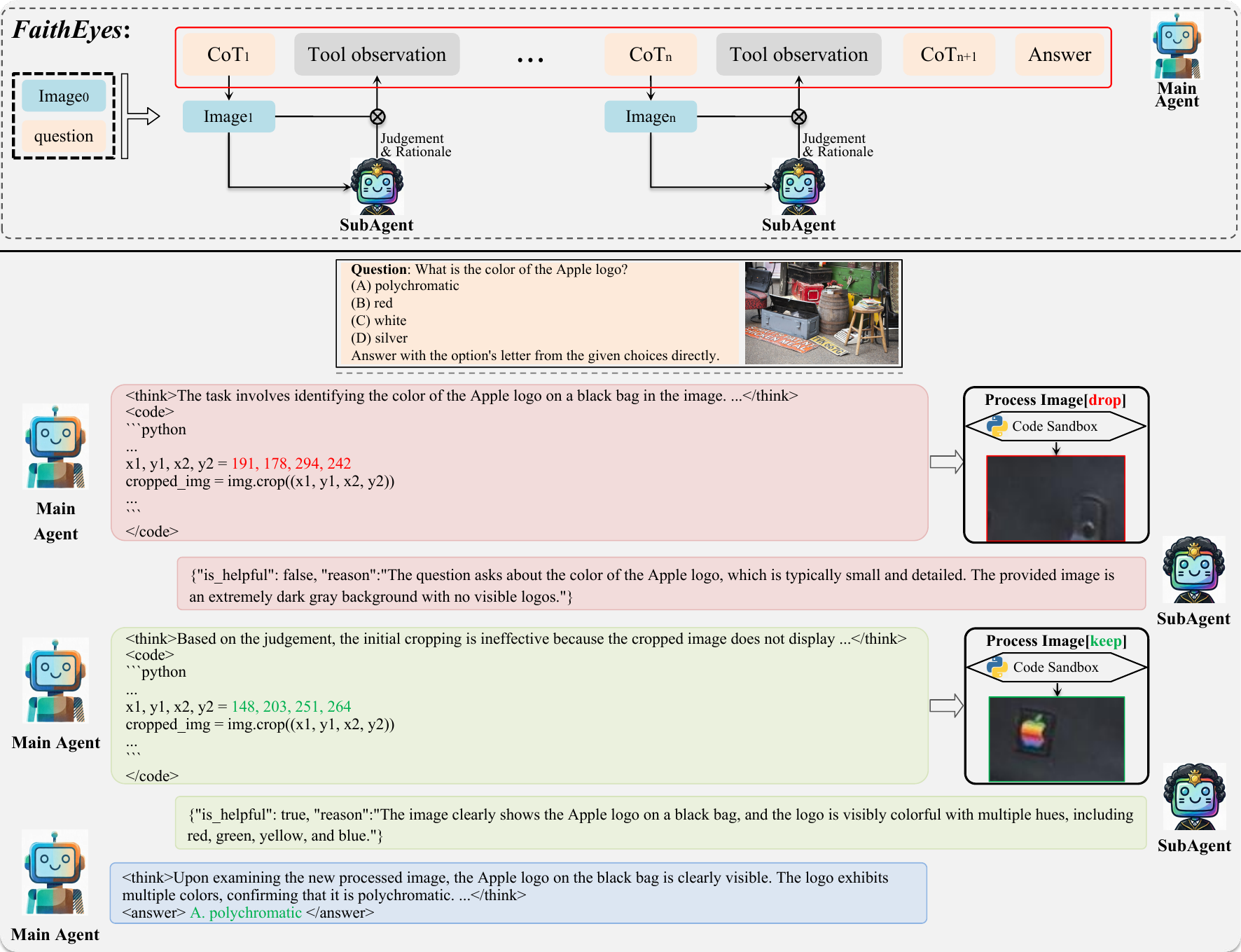}
    \caption{Illustration of our \emph{\textbf{FaithEyes}} framework. The main agent calls tools to obtain the process images, and the subagent gives the judgement to help further reasoning.}
    \label{framework}
    \vspace{-5mm}
\end{figure}

This low process-image faithfulness could be attributed to two coupled shortages. Firstly, existing reward designs fail to distinguish useful from useless tool calls. Identical rewards are assigned whenever the final answer is correct and a tool is invoked, regardless of whether the process image actually aids the solution~\citep{zheng2025deepeyes,yang2025thinking}. Secondly, tool feedback provides only the resultant image, without any indication of its helpfulness~\citep{hou2026codev,zhang2025thyme}. The model lacks explicit incentive to examine whether intermediate visual evidence is relevant, particularly when it can take a shortcut to the answers using prior knowledge or random guessing. Over training, the model progressively learns to output decorative tool calls and invoke tools to harvest the bonus while never engaging with their outputs, which wastes inference cost on unnecessary operations and fundamentally limits capacity for genuinely demanding visual reasoning. To address these issues, we introduce \emph{\textbf{FaithEyes}}, a multi-agent self-judging framework for agentic VLMs. Concretely, a VLM judges each process image's helpfulness for answering the question and provides the corresponding rationale. The process image, together with this judgment, is incorporated into the context as the tool observation, providing an explicit clue for further reasoning. Note that, we discard the process images judged as unhelpful and return only their judgment as the tool observation, while the process images judged as helpful are returned alongside their judgment. This strategy can effectively avoid unnecessary computation and interference from unhelpful images. Concurrently, the judgment result is also used to compute a process-level tool reward scaled by a helpful-tool ratio, i.e., the proportion of successfully executed and genuinely helpful tool calls, which can effectively prevent reward hacking. To avoid external model dependency during evaluation, we design a multi-agent framework where the model itself serves as a subagent to judge the main agent's tool-call helpfulness, thereby ensuring the availability of judgment. The overall framework and an interaction trajectory are illustrated in Figure~\ref{framework}.

Our overall training pipeline comprises two stages: supervised fine-tuning (SFT) and reinforcement learning (RL). The SFT stage is responsible for cold-start initialization, equipping the model with three essential capabilities prior to the RL stage: (1) tool invocation—the ability to write executable code for obtaining auxiliary visual information; (2) judgment—the ability to judge whether a process image is helpful for answering the question and give the rationale behind its assessment; and (3) interactive reasoning—the ability to adapt subsequent reasoning based on the tool observation (including process images and judgment) to arrive at correct answers. The RL stage further enhances generalization and robustness. We use the GRPO algorithm \citep{shao2024deepseekmath} for RL training, combining four reward types: accuracy reward, consistency reward \citep{team2025kwai,zhang2025r1}, format reward, and tool reward. For the tool reward, we use the proportion of helpful tool calls to prevent the reward hacking for tool usage. All SFT and RL datasets are derived from the same public sources used by our baselines~\citep{zheng2025deepeyes,zhang2025thyme}. We primarily evaluate visual perception (i.e., V$^{*}$ Bench~\citep{wu2024v}, HR-Bench 4K and 8K \citep{wang2025hrbench}) and visual reasoning (i.e., MathVista \citep{lu2023mathvista}, MathVerse \citep{zhang2024mathverse}, and MathVision \citep{wang2024measuring}) benchmarks. FaithEyes achieves comparable performance with previous state-of-the-art models while substantially improving tool faithfulness. Moreover, FaithEyes also reduces inference cost, consuming obviously fewer tokens than prior agentic VLMs.

Overall, our contributions can be summarized as follows:
\begin{itemize}
\item We attribute the unfaithful tool use of agentic VLMs to two coupled shortages (i.e., the tool reward that fails to distinguish useful from useless calls and the tool feedback that carries no signal of usefulness). The former shortage stems from the reward design itself, and we validate the latter one with quantitative analysis.
\item We introduce \emph{\textbf{FaithEyes}}, a multi-agent self-judging framework in which the model serves as its own subagent to judge each process image, and the resulting judgement is both injected into the tool observation to help reasoning and used to calculate the tool reward via a helpful-tool ratio, jointly addressing the above two shortages.
\item Training with a SFT + RL pipeline on adapted open-source data, FaithEyes attains competitive accuracy across visual perception and reasoning benchmarks, and markedly improves process image faithfulness and reduces inference cost.
\end{itemize}

\section{Related works}
\paragraph{Agentic vision-language models.}
Agentic VLMs extend multi-modal reasoning beyond a single forward pass by interleaving textual reasoning with tool calls~\citep{su2025thinking}. Rather than treating the image as static context, these models actively decide when and how to invoke tools (e.g., cropping and zooming, executing code) and incorporate the returned results into subsequent reasoning~\citep{su2026pixel,deepeyesv2}. By grounding intermediate conclusions on verifiable tool outputs, this paradigm improves accuracy and supports more interpretable problem solving. To elicit this ability, some works apply prompt engineering to top-tier VLMs~\citep{hu2024visual,lee2025interactive,fu2025refocus}, while others use supervised fine-tuning~\citep{ge2025advancing,zhao2025pyvision} or reinforcement learning~\citep{lai2025mini,zheng2025deepeyes,zhang2025thyme} on affordable small VLMs. Some other works~\citep{xu2025visual,chern2025thinking,han2025controlthinker,jiang2026t2i,shi2026mathcanvas} use image generation to achieve latent "thinking with images". In this work, we focus on reinforcement-learning-based methods with explicit tool calls that represents the state-of-the-art.

\paragraph{Faithfulness of tool use in agentic VLMs.}
Despite strong benchmark performance, a growing body of work has revealed that agentic VLMs often use visual tools unfaithfully. \citet{liu2025faithful} probes the faithfulness of multi-modal chain-of-thought through intervention, showing that predictions remain nearly unchanged when visual thoughts are corrupted. This indicates that the intermediate visual evidence is largely ignored. Similar observations are reported~\citep{yang2026position}, which find that the model can obtain nearly the same performance without process images and the answer tokens focus on the initial image more than process images. These failures are commonly traced to reward designs that encourage the mere presence of tool calls rather than their usefulness, leaving room for reward hacking. CodeV~\citep{hou2026codev} rethinks reward from a process perspective and uses a judge model to assign step-level rewards to each tool output. However, CodeV only uses the judgement as a reward signal. In contrast, we additionally feed the judgement back into the reasoning context as part of the tool observation, so that it actively helps subsequent reasoning, and we further keep this mechanism at inference through a multi-agent framework.

\paragraph{Multi-agent systems.}
Multi-agent systems coordinate multiple LLM-based agents with specialized roles to solve tasks beyond a single agent~\citep{tran2025multi,xi2025rise}. They vary in topology from centralized orchestration that decomposes and assigns subtasks~\citep{qian2023chatdev,wu2023autogen,du2025survey}, to decentralized peer interaction~\citep{wang2025large}. They also vary in interaction patterns from cooperative collaboration where a recurring design assigns one agent an evaluative role to critique or verify others~\citep{chen2024agentverse}, to large-scale social simulation~\citep{park2023generative,zhao2025llm}. Recent work begins to integrate this paradigm with agentic VLMs. SCoT~\citep{yang2025thinking} lets a main agent decompose a complex visual query into atomic subtasks that are solved by parameter-sharing subagent, reformulating interleaved multi-modal reasoning as a language-only chain-of-thought. While we likewise adopt a self-calling design, our objective differs fundamentally. Concretely, we employ the subagent to judge the helpfulness of each process image produced by the main agent. This judgement is consumed both as contextual feedback and as a reward signal. In this sense, FaithEyes also resonates with recursive self-improvement~\citep{chen2026recursive}, where our self-judging design uses the model's own assessment of its tool outputs for a reward signal, with no dependence on external evaluator.

\section{Method}
\subsection{Preliminaries and analysis}\label{sec:prelim}
\paragraph{Agentic VLM reasoning.}
Many visual questions can be answered more reliably when a model actively interrogates the image rather than committing to a single holistic glance. The \emph{thinking-with-images} paradigm~\citep{zheng2025deepeyes,su2025thinking} builds on this intuition by treating the image not as a static input, but as a dynamic and manipulable cognitive workspace. The model interleaves textual reasoning with tool calls which aim to acquire helpful visual evidence. Formally, given a visual question $\mathbf{x}=(I,Q)$ comprising an image $I$ and a textual query $Q$, the reasoning trajectory of a policy model $\pi_\theta$ can be formulated as
\begin{equation}
\tau = (\mathbf{x}, a_1, o_1, a_2, o_2, \ldots, a_T),\quad a_t \sim \pi_\theta(\cdot \mid \mathbf{x}, a_{<t}, o_{<t})
\end{equation}
where each action $a_t$ is generated by the policy model conditioned on the input question $\mathbf{x}$ and the interaction history $\{a_{<t}, o_{<t}\}$, and $o_t$ denotes the observation returned by executing action $a_t$. Each action $a_t$ couples a thinking process with the concrete move. Concretely, the model first reasons in free-form text and then commits to one of two moves: either emitting a final answer that terminates the trajectory $\tau$, or invoking a tool (e.g., cropping, zooming, coding, or other operations) to obtain additional information. The tool observation $o_t$ is the feedback obtained by executing the tool, and it is governed by the environment and action rather than modeled by the policy $\pi_\theta$.

\paragraph{Tool faithfulness problem.}
A successful tool call $a_t$ produces process images $I_t$ together with optional logs or numeric outputs. These are returned as the observation $o_t$ and appended to the context, thereby serving as additional visual evidence for subsequent reasoning. The process image collection $\{I_t\}$ constitutes the intermediate visual evidence gathered by the model during reasoning, and ideally one tool call should return the process images that actually capture the evidence the question asks about. In practice, however, recent studies show that agentic VLMs use tools unfaithfully. Concretely, even when the answer is correct, only about half of the samples actually contain at least one tool call with the queried target~\citep{hou2026codev}, so many process images are decorative or misaligned. The tools are invoked, but their outputs are useless, which wastes inference cost without strengthening visual reasoning. Here we conjecture two coupled shortages. 1) Undifferentiated tool reward: the tool reward is granted whenever the answer is correct and a tool is invoked~\citep{zheng2025deepeyes}, regardless of whether any $I_t$ contributes. Helpful and unhelpful calls receive identical credit, and such usage-centric rewards let hacking persist. A bare tool-call bonus only inflates call frequency without improving how outputs are helpful. 2) Usefulness-agnostic feedback: the observation $o_t$ contains the process images, but by itself it carries no explicit signal of that image's usefulness, so the model is never prompted to examine whether the retrieved evidence is relevant. Concretely, the first cause is entailed by the reward design itself. Many representative methods~\citep{zheng2025deepeyes,yang2025thinking} grant the tool bonus purely on the co-occurrence of a correct answer and a tool call, while others~\citep{zhang2025thyme,deepeyesv2} forgo the tool reward, leaving tool behavior entirely unguided. The second cause concerns the inference-time context and is less self-evident, so we probe it with quantitative analysis and ask whether the missing usefulness signal is important. We try to directly inject the judgement signals into the tool observation of DeepEyes~\citep{zheng2025deepeyes} and Thyme~\citep{zhang2025thyme} to help their reasoning without any retraining. Here we synthesize this signal (i.e., a binary helpful/unhelpful verdict with a short rationale) by requesting Qwen3-VL-235B-A22B-Instruct~\citep{bai2025qwen3} with Prompt~\ref{tmpl:sub_prompt}. The results are provided in Table~\ref{tab:with_judgement}. As we can see, injecting judgement signals improves accuracy on most benchmarks for both DeepEyes and Thyme, indicating that the judgement supplies decision-relevant information the model previously failed to extract from the process image alone. However, the gains are not uniform and DeepEyes regresses on HR-Bench, suggesting that a judgement produced by an external model and appended without any retraining is not always well calibrated to the frozen policy. We need to learn the judging and reasoning behaviors jointly, and this motivates our FaithEyes.

\begin{table*}[t]
\centering
\caption{Performance changes of agentic VLMs (i.e., DeepEyes and Thyme) when the judgement of process images is added to tool observation.}
\scalebox{0.8}{
\begin{tabular}{l|ccc|ccc}
\toprule
 & \multicolumn{3}{c|}{\textbf{Perception}} & \multicolumn{3}{c}{\textbf{Reasoning}} \\
Model & V$^{*}$ & HR-Bench 4K & HR-Bench 8K & MathVista & MathVerse & MathVision \\
\midrule
DeepEyes              & 84.3 & 74.2 & 70.4 & 68.7 & 44.3 & 28.3 \\
DeepEyes w/ Judgement & 86.2$_{\textcolor{mygreen}{\uparrow 1.9}}$ & 73.9$_{\textcolor{red!50}{\downarrow 0.3}}$ & 68.1$_{\textcolor{red!50}{\downarrow 2.3}}$ & 69.3$_{\textcolor{mygreen}{\uparrow 0.6}}$ & 46.8$_{\textcolor{mygreen}{\uparrow 2.5}}$ & 28.6$_{\textcolor{mygreen}{\uparrow 0.3}}$ \\
\midrule
Thyme                 & 82.7 & 74.6 & 69.6 & 69.9 & 44.4 & 28.6 \\
Thyme w/ Judgement    & 85.8$_{\textcolor{mygreen}{\uparrow 3.1}}$ & 75.5$_{\textcolor{mygreen}{\uparrow 0.9}}$ & 70.7$_{\textcolor{mygreen}{\uparrow 1.1}}$ & 71.4$_{\textcolor{mygreen}{\uparrow 1.5}}$ & 46.1$_{\textcolor{mygreen}{\uparrow 1.7}}$ & 29.6$_{\textcolor{mygreen}{\uparrow 1.0}}$ \\
\bottomrule
\end{tabular}}
\label{tab:with_judgement}
\vspace{-3mm}
\end{table*}

\subsection{\emph{FaithEyes}: multi-agent self-judging framework}\label{sec:faithEyes}
FaithEyes targets the above two shortages jointly: it injects an explicit usefulness signal into the tool observation and reuses it to differentiate the tool reward. The framework is a single self-calling VLM, playing two roles: a \emph{main agent} that solves the question and a \emph{subagent} that judges each process image the main agent produces. We use executable code as the tool interface~\citep{zhang2025thyme,hou2026codev} which subsumes cropping, zooming, rotation, contrast adjustment, and arithmetic within one expressive channel and lets the model compose arbitrary operations rather than pick from a fixed set. Figure~\ref{framework} illustrates the whole framework and an interaction trajectory.

\paragraph{Main agent.} Conditioned on $\mathbf{x}=(I,Q)$ and history $\{a_{<t},o_{<t}\}$, the main agent generates $a_t\sim\pi_\theta(\cdot\mid\mathbf{x},a_{<t},o_{<t})$, interleaving free-form reasoning with either a final answer (terminating $\tau$) or a tool call. When a tool is invoked, the emitted code block is regex-extracted and run in a Python sandbox environment with read-only access to $I$. Execution returns a raw observation $\tilde{o}_t$, which may contain one or more process images $I_t$, numerical calculation results, extracted OCR results, and so on. The system and user prompts of main agent are provided in Prompt~\ref{tmpl:main_prompt}.

\paragraph{Subagent.} A process image is only useful if it actually exposes the information the question asks for, otherwise it is superfluous context and even a vehicle for reward hacking. FaithEyes therefore assigns a dedicated subagent, realized by the same model under a separate subagent prompt, to evaluate $\{I_t\}$. Concretely, the subagent maps the process image and the original question $Q$ to a structured verdict $(h_t,e_t)=g_\theta(\cdot\mid Q,\{I_t\})$ where $h_t\in\{\mathrm{True},\mathrm{False}\}$ is a binary helpfulness label, $e_t$ is a free-form rationale, and $g_\theta$ denotes the policy under the subagent prompt, as shown in Prompt~\ref{tmpl:sub_prompt}. Before judging, we downscale each process image $I_t$ to at most 1K resolution to reduce the judging burden if necessary. The judging rubric sets $h_t=\mathrm{True}$ when the object or attribute queried by $Q$ is visible in $\{I_t\}$ so that the target and its relevant attributes (e.g., color, position, shape) can be identified, and $h_t=\mathrm{False}$ when the queried target is absent. The verdict is emitted as a single JSON line, \texttt{\{``is\_helpful'': $h_t$, ``reasons'': $e_t$\}}, keeping the supervision both machine-parseable and human-readable. Two design choices are worth emphasizing. First, the subagent conditions only on $(Q, \{I_t\})$ and never on the main agent's private chain-of-thought or code. This evidence-centric view is far cheaper and more stable to assess than diagnosing internal reasoning, and it generalizes to arbitrary tool outputs without dense annotations such as ground-truth boxes. Second, the subagent is instantiated by the model itself rather than an external judge model. This keeps the judgment available at inference and needs no external model for a fair comparison.

\paragraph{Verdict-guided tool observation.} The verdict $(h_t,e_t)$ is folded back into the reasoning context as part of the observation, so that the main agent is explicitly told whether and why the process images help or not. When $h_t=\mathrm{True}$, we return both $\{I_t\}$ and $(h_t,e_t)$, and when $h_t=\mathrm{False}$, we discard $\{I_t\}$ and return only $(h_t,e_t)$. This asymmetry serves two purposes. Firstly, by telling the main agent why a crop misses the target, the negative verdict pushes it to output a corrected tool call that targets the right region, or to fall back to reasoning over the original image $I$ when the queried evidence is genuinely unavailable from any crop, rather than silently proceeding on irrelevant evidence. Here the model acts on an explicit verdict and either repairs the crop or makes a justified fallback. This directly remedies the usefulness-agnostic feedback identified above. Second, removing unhelpful process images from the context both curtails interference from spurious visual content and substantially reduces the visual token and computation cost of invalid tool calls, since the process images dominate the inference overhead of agentic VLMs. Together with the reward-side use of $h_t$ introduced below, the same $h_t$ serves a dual role: it scales the tool reward and steers the next reasoning step.

\subsection{Training pipeline}\label{sec:training}
Following common practice~\citep{zhang2025thyme,hou2026codev}, we adopt a two-stage pipeline: a cold-start supervised fine-tuning (SFT) stage to elicit the expected capabilities, followed by a reinforcement learning (RL) stage that reinforces them.

\paragraph{Cold-start SFT.}
The SFT stage cold-starts three capabilities prior to the RL stage. (i) Code-based problem solving ability. Some base VLMs like Qwen2.5-VL-7B-Instruct~\citep{bai2025qwen25vltechnicalreport} do not natively write code to solve visual problems~\citep{zhang2025thyme,deepeyesv2}, so this ability needs to be bootstrapped by SFT. (ii) Faithfulness judging ability. The model needs to judge whether each process image helps answer the question and emit the verdict as JSON. While the base model itself typically has judging capabilities, it's necessary to align some preferences regarding clarity and crop concentration to avoid judging large lazy cropping as helpful or judging precise cropping of small objects as unhelpful due to their low clarity. Some images of small objects have low native resolution, which is not a tool error. (iii) Feedback-driven reasoning ability. The model needs to condition subsequent reasoning on the subagent's judgement rather than ignore it.

\paragraph{Reinforcement learning with GRPO.} We further reinforce generalization with Group Relative Policy Optimization~\citep{shao2024deepseekmath}. For each $\mathbf{x}$, we sample $G$ trajectories $\{\tau_i\}_{i=1}^G\sim\pi_\theta(\cdot\mid\mathbf{x})$ with rewards $r_i=r(\tau_i)$ and the group-normalized advantage $A_{i,j}=(r_i-\mathrm{mean}(\{r_k\}_{k=1}^G))/\mathrm{std}(\{r_k\}_{k=1}^G)$ broadcast over all generated tokens. The policy maximizes the clipped objective with a Kullback-Leibler divergence trust region to the cold-start reference $\pi_{\mathrm{ref}}$, and the loss function is
\begin{equation}
\scalebox{0.82}{$\mathcal{J}(\theta)=\mathbb{E}_{\mathbf{x}\sim\mathcal{D},\{\tau_i\}\sim\pi_{\theta}(\cdot\mid\mathbf{x})}\!\left[\frac{1}{\sum_{i=1}^G|\tau_i|}\sum_{i=1}^{G}\sum_{j=1}^{|\tau_i|}\!\Big(\min\!\big(\rho_{i,j}A_{i,j},\,\mathrm{clip}(\rho_{i,j},1{-}\epsilon,1{+}\epsilon)A_{i,j}\big)-\beta\,\mathbb{D}_{\mathrm{KL}}[\pi_\theta\!\parallel\!\pi_{\mathrm{ref}}]\Big)\right]$}
\end{equation}
where $\rho_{i,j}=\pi_\theta(a_{i,j}\mid s_{i,j})/\pi_{\mathrm{ref}}(a_{i,j}\mid s_{i,j})$ and $s_{i,j}$ is the trajectory history up to token $j$. Note that the tool observations (including process images and judgement) are excluded from token counts and advantage estimation, since they are environmental feedback and do not depend on the policy model.

\paragraph{Reward design.} In this work, our trajectory reward $r(\tau)$ contains 4 types of rewards, i.e.,
\begin{equation}\label{reward}
    r(\tau)=r_{\mathrm{acc}}+\lambda_{\mathrm{fmt}}\cdot r_{\mathrm{fmt}}+\lambda_{\mathrm{cons}}\cdot r_{\mathrm{cons}}+\lambda_{\mathrm{tool}}\cdot r_{\mathrm{tool}}
\end{equation}
The first three rewards bookkeep answer correctness and surface quality, while the tool reward targets faithful tool calls. We detail each reward below:\\
1) \emph{Accuracy reward} $r_{\mathrm{acc}}\in\{0,1\}$ measures whether the extracted final answer matches the ground truth. Since the answers in our dataset are not always numeric or formulaic, we first attempt rule-based exact or programmatic matching, and fall back to a Qwen2.5-VL-72B-Instruct LLM-as-judge that assesses semantic equivalence against the reference answer whenever the rule-based matcher fails.\\
2) \emph{Format reward} $r_{\mathrm{fmt}}\in\{0,-1\}$ enforces the prescribed output structure. The main agent is required to wrap its reasoning, code, and conclusion within the \texttt{\textless think\textgreater}, \texttt{\textless code\textgreater}, and \texttt{\textless answer\textgreater} tags respectively, so that each component can be reliably parsed, especially the executable code. A trajectory receives $r_{\mathrm{fmt}}=-1$ whenever this structure is malformed improperly, and $0$ otherwise.\\
3) \emph{Consistency reward} $r_{\mathrm{cons}}\in\{0,-1\}$~\citep{team2025kwai,zhang2025r1} examines whether the final answer is logically entailed by the preceding reasoning, rather than appended as an unsupported guess. Concretely, we feed the trailing segment of the reasoning together with the answer to Qwen2.5-VL-72B-Instruct, which judges whether the conclusion follows from the stated argument.\\
4) \emph{Tool reward} $r_{\mathrm{tool}}\in[0,1]$ is the faithfulness-targeting term. A flat bonus granted on any tool call~\citep{zheng2025deepeyes} rewards the mere presence of a call over its usefulness, inflating call frequency without improving how the returned evidence is useful. We instead scale the bonus by the fraction of tools that are both executable and genuinely helpful, i.e.,
\begin{equation}
    r_{\mathrm{tool}}=1-\frac{n_{\mathrm{fail}}+n_{\mathrm{unhelpful}}}{n_{\mathrm{tool}}},\quad \text{if  } n_{\mathrm{tool}}>0
\end{equation}
It aggregates over all tools in a trajectory: $n_{\mathrm{tool}}$ counts all tool calls, $n_{\mathrm{fail}}$ counts those that fail to execute or yield no output, and $n_{\mathrm{unhelpful}}$ counts those judged as unhelpful by the subagent. This helpfulness gate ensures that decorative crops which happen to co-occur with a lucky answer still earn nothing, since they drive $r_{\mathrm{tool}}$ toward $0$. Since $r_{\mathrm{tool}}$ is built from the very same $h_t$ that guides further reasoning in~\S\ref{sec:faithEyes}, the model is pushed to call faithful tools through both observation and reward. Unlike prior works~\citep{zheng2025deepeyes}, we do not gate the tool reward on answer correctness, since this leads to a large fraction (up to 18$\%$) of non-executable code during RL process and ultimately degenerates the policy into avoiding tool use altogether, as shown in Figure~\ref{code_failure}.

\section{Experiments}
\subsection{Experimental setup}\label{sec:exp:setup}
\paragraph{Data preparation.}
To cold-start the desired three capabilities (i.e., code writing, judgement and interactive reasoning) in SFT stage, we adapt the open-source SFT data from Thyme~\citep{zhang2025thyme}, which includes single- and two-tool-call trajectories. Their structure gives natural ground-truth judgement: the sole call in a single-tool-call trajectory is labeled as $h_t=\mathrm{True}$, and in a two-tool-call trajectory, the first call is labeled as $h_t=\mathrm{False}$ and the second one is labeled as $h_t=\mathrm{True}$. We prompt Qwen3-VL-235B-A22B-Instruct~\citep{bai2025qwen3} to write the matching rationale $e_t$ for each label (Prompt~\ref{tmpl:sub_prompt}) and compose observations as in FaithEyes: helpful images return with their judgement, while unhelpful ones are dropped and only their judgements are returned. We further use Qwen3-VL-235B-A22B-Instruct with Prompt~\ref{tmpl:rewrite_prompt_helpful} and Prompt~\ref{tmpl:rewrite_prompt_unhelpful} to rewrite the model reasoning following the judgment, ensuring that subsequent reasoning relies on the preceding judgment. Finally, we construct two interleaved supervision sources: problem-solving trajectories for the main agent and judgment trajectories for the subagent. We use all the problem-solving trajectories and half of the judgment trajectories, and finally obtain 457K SFT data. For RL stage, we collect the RL data from Thyme-55K~\citep{zhang2025thyme} and DeepEyes-47K~\citep{zheng2025deepeyes}, and filter out the questions that our SFT model can directly answer with 100$\%$ accuracy across 8 inference attempts. All SFT and RL datasets are derived from the same public sources used by our baselines.

\paragraph{Implementation details.}
We select Qwen2.5-VL-7B-Instruct and Qwen3-VL-8B-Instruct as the initial models. For the SFT stage, we finetune Qwen2.5-VL-7B-Instruct for 3 epochs and Qwen3-VL-8B-Instruct for 5 epochs with a learning rate of $1e^{-5}$ and batch size of 128. For the RL stage, we adopt GRPO~\citep{shao2024deepseekmath} with batch size of 128, 12 rollouts per sample and a learning rate of $1e^{-6}$. We set the reward coefficients in Equation~\ref{reward} as $\lambda_{\mathrm{fmt}}=0.2$, $\lambda_{\mathrm{cons}}=0.2$ and $\lambda_{\mathrm{tool}}=0.2$, so that answer correctness remains the dominant signal and the tool term acts as a faithfulness regularizer. More data and training details are provided in the appendix.

\begin{table*}[t]
\centering
\caption{\textbf{Performance on visual perception and reasoning benchmarks.} The best result of each benchmark in agentic VLMs is \textbf{bolded} and the second best is \underline{underlined}.}
\scalebox{0.77}{
\begin{tabular}{llc|ccc|ccc}
\toprule
 & & & \multicolumn{3}{c|}{\textbf{Perception}} & \multicolumn{3}{c}{\textbf{Reasoning}} \\
Models & Tools & Size & V$^{*}$ & HR-Bench 4K & HR-Bench 8K & MathVista & MathVerse & MathVision \\
\midrule
\multicolumn{9}{c}{\textit{Proprietary Models}} \\
\midrule
GPT-4o & - & - & 64.4 & 63.1 & 61.3 & 63.7 & 35.3 & 35.9 \\
\midrule
\multicolumn{9}{c}{\textit{VLM w/o Tools}} \\
\midrule
Qwen2.5-VL & - & 7B & 75.0 & 68.6 & 63.6 & 67.9 & 45.5 & 21.4 \\
Qwen2.5-VL & - & 32B & 87.9 & 73.9 & 70.4 & 72.2 & 40.0 & 35.2 \\
Qwen3-VL & - & 8B & 86.4 & 78.9 & 74.6 & 77.2 & 62.1 & 53.9 \\
\midrule
\multicolumn{9}{c}{\textit{Agentic VLM (Qwen2.5-VL-7B)}} \\
\midrule
DeepEyes & Crop & 7B & 84.3 & 74.2 & 70.4 & 68.7 & 44.3 & 28.3 \\
Pixel-Reasoner & Crop & 7B & 84.3 & 74.0 & 66.9 & 71.2 & 46.9 & 26.3 \\
Thyme & Code & 7B & 82.7 & 74.6 & 69.6 & 69.9 & 44.4 & 28.6 \\
CodeV & Code & 7B & 84.8 & 76.1 & 71.3 & \underline{71.8} & 49.2 & \textbf{33.6} \\
PyVision-Image & Code & 7B & \textbf{88.7} & \textbf{78.1} & \textbf{74.3} & - & \textbf{55.8} & 28.7 \\
\textbf{FaithEyes (Ours)} & Code & 7B & \underline{87.4} & \underline{77.8} & \underline{72.9} & \textbf{73.1} & \underline{51.0} & \underline{29.9} \\
\midrule
\multicolumn{9}{c}{\textit{Agentic VLM (Qwen3-VL-8B)}} \\
\midrule
BEE         & Latent & 8B & 90.6 & 80.2 & 76.8 & - & - & - \\
\textbf{FaithEyes (Ours)} & Code & 8B & 89.9 & 83.2 & 78.5 & 78.7 & 67.6 & 46.1 \\
\bottomrule
\end{tabular}}
\label{tab:main_results}
\vspace{-3mm}
\end{table*}

\begin{figure}[t]
    \centering
    \includegraphics[width=\linewidth]{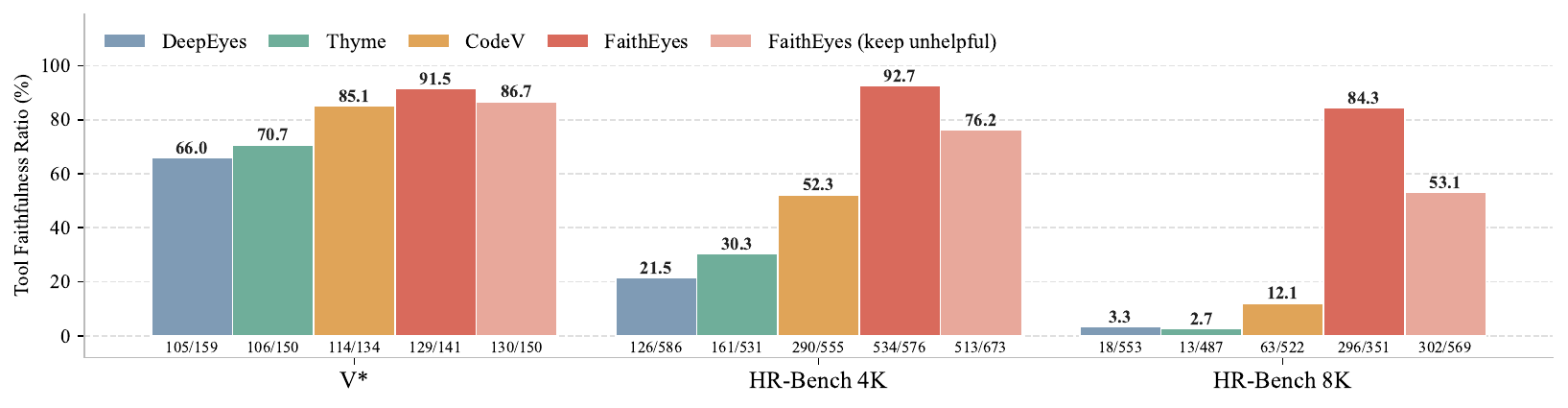}
    \caption{Tool faithfulness ratio on V$^{*}$, HR-Bench 4K and HR-Bench 8K. The bottom of the bar shows the number of trajectories with faithful tools and the total trajectories.}
    \label{faithfullness_bar}
    \vspace{-5mm}
\end{figure}

\subsection{Main results}\label{sec:exp:main}
\paragraph{Accuracy and faithfulness.}
In this work, we mainly examine visual perception (i.e., V$^{*}$, HR-Bench 4K and HR-Bench 8K) and reasoning (i.e., MathVista, MathVerse and MathVision) benchmarks. Table~\ref{tab:main_results} compares FaithEyes against the proprietary model (GPT-4o~\citep{hurst2024gpt}), tool-free open-source VLMs (Qwen2.5-VL-7B/32B, Qwen3-VL-8B), and state-of-the-art agentic VLMs (DeepEyes, Pixel-Reasoner~\citep{su2026pixel}, Thyme, CodeV, PyVision-Image~\citep{zhao2026pyvision}, and BEE~\citep{chen2026beyond}). To ensure robustness, we report the Avg@10 results. FaithEyes achieves comparable accuracy with the state-of-the-art baseline PyVision-Image, which is trained with a far broader SFT and RL data mixture. The margin over the data-comparable baselines (i.e., DeepEyes, Thyme, CodeV) is largest on the small-object, high-resolution perception benchmarks, which require a clear look at the small correct region, and this is consistent with tool calls more reliably landing on it. For Qwen3-VL-8B, FaithEyes also achieves significant improvements over the base model, except on MathVision. We further examine the tool faithfulness ratio of problem-solving trajectories. Concretely, we only consider the correct-answer trajectories with process images and use Qwen3-VL-235B-A22B to determine whether the trajectory contains process images that are helpful for answering the question, and the judgement prompt is Prompt~\ref{tmpl:jud_prompt}. Here we use a prompt distinct from training to ensure a fair comparison. As shown in Figure~\ref{faithfullness_bar}, FaithEyes obtains a substantially higher tool faithfulness ratio than our agentic baselines. The tool calls of previous methods often produce many irrelevant process images. FaithEyes not only effectively reduces unnecessary process images, but also substantially increases the number of helpful ones, as shown on HR-Bench 8K. Furthermore, one may worry that this improvement is an artifact of the automatic filtering of FaithEyes, so we also evaluate FaithEyes in the \emph{keep-unhelpful} trajectories where we do not discard any images judged as unhelpful. Even without dropping any self-judged-unhelpful images, FaithEyes (keep unhelpful) still outperforms the strongest baseline (i.e., CodeV) by $1.6\sim41$ points across three benchmarks. This confirms that the faithfulness gains of FaithEyes are not merely a byproduct of filtering, but reflect a genuinely more faithful tool-use policy that produces the process images actually helpful.

\begin{table}[t]
\centering
\caption{Inference cost comparison. \emph{Response tokens} is the average token number of responses and tool observations, and \emph{Tool calls} is the average number of tool invocations per sample. For FaithEyes, we also include the subagent's prompt and response in the total token count.}
\scalebox{0.83}{
\begin{tabular}{lcccccc}
\toprule
& \multicolumn{2}{c}{V$^*$} & \multicolumn{2}{c}{HR-Bench 4K} & \multicolumn{2}{c}{HR-Bench 8K} \\
\cmidrule(lr){2-3}\cmidrule(lr){4-5}\cmidrule(lr){6-7}
Models & Response tokens & Tool calls & Response tokens & Tool calls & Response tokens & Tool calls \\
\midrule
DeepEyes   & 4726 & 1.0 & 15499 & 1.0 & 16514 & 1.0 \\
Thyme      & 5339 & 1.0 & 15328 & 1.0 & 17955 & 1.1 \\
CodeV      & 5340 & 1.3 & 22505 & 3.2 & 31088 & 3.1 \\
\midrule
FaithEyes  & 4020 & 0.9 & 12580 & 0.9 & 14269 & 0.8 \\
\bottomrule
\end{tabular}}
\label{tab:inference_cost}
\vspace{-5mm}
\end{table}

\paragraph{Training and inference cost.}
A natural concern is that the additional self-judging step introduces extra overhead. However, FaithEyes is no more expensive to train and cheaper at inference than the baselines. For training, the prior agentic VLMs rely on external large models for judging. CodeV deploys Qwen2.5-VL-32B to score every tool output, and Thyme and DeepEyes use Qwen2.5-VL-72B for result and consistency judging. FaithEyes with the self-judging of a 7B VLM is cheaper than CodeV and incurs small additional cost compared to DeepEyes and Thyme. For inference, we report the total response tokens and tool call number as the inference cost in Table~\ref{tab:inference_cost}. FaithEyes consumes fewer tokens on all three benchmarks, especially for the high-resolution ones. Compared with CodeV, FaithEyes can save 24.7$\%$, 44.1$\%$ and 54.1$\%$ tokens on V$^*$, HR-Bench 4K and HR-Bench 8K respectively. In fact, each self-judging step consumes only $\sim$1.5k tokens on average, far less than a self-reflection turn that must re-attend to all previous reasoning and tool observations. But it lets the main agent eliminate multi-round re-cropping (i.e., the fewest tool calls) and discard unhelpful process images rather than carrying them across each subsequent turn. Since a process image, especially a poorly targeted crop of a high-resolution input, occupies a large number of vision tokens, higher faithfulness brings more precise cropping and thereby wastes fewer tokens.

\subsection{Ablation studies}
\begin{table*}[t]
\centering
\caption{Model design ablation: average accuracy and tool faithfulness ratio across perception and reasoning benchmarks. Here we report the ratio of \emph{keep-unhelpful} protocol for clear comparison.}
\scalebox{0.88}{
\begin{tabular}{l|cc|ccc}
\toprule
             & \multicolumn{2}{c|}{\textbf{Accuracy}} & \multicolumn{3}{c}{\textbf{Tool Faithfulness Ratio}} \\
Model Design & Perception & Reasoning  & V$^{*}$ & HR-Bench 4K & HR-Bench 8K \\
\midrule
FaithEyes                     & 79.4 & 51.3 & 86.7 & 76.2 & 53.1 \\
\midrule
w/o Judgement injection       & 77.8 & 49.5 & 78.8 & 57.0 & 15.7 \\
w/o Reward scaling            & 78.1 & 50.2 & 75.5 & 40.8 & 9.8 \\
w/o Both                      & 76.2 & 48.4 & 72.2 & 33.5 & 4.6 \\
\midrule
w/ Qwen3-VL-235B-A22B judge   & 79.7 & 51.2  & 88.3 & 78.6 & 56.2 \\
\bottomrule
\end{tabular}}
\label{tab:design_ablation}
\vspace{-3mm}
\end{table*}
\paragraph{Model design ablation.}
Table~\ref{tab:design_ablation} ablates the two mechanisms using the same judgement signal (i.e., injecting the verdict into the tool observation (\emph{judgement injection}) and scaling the tool reward by the helpful-tool ratio (\emph{reward scaling})), and reports accuracy and the tool faithfulness ratio of keep-unhelpful trajectories for clear comparison. Firstly, removing either mechanism degrades accuracy and removing judgement injection hurts most, where the main agent loses the explicit signal of whether the retrieved evidence is trustworthy, so it struggles to recognize and re-crop off-target process images and instead proceeds on unreliable evidence. Secondly, removing either mechanism also sharply collapses tool faithfulness, but here the ordering flips and removing \emph{reward scaling} hurts most. In fact, reward scaling provides the incentive that drives the model to produce helpful crops, while judgement injection provides the scaffold that lets the model recognize and act on them. The incentive is unactionable without the scaffold, and the scaffold is ignored without the incentive. Removing both is weakest on all metrics with the largest accuracy and tool faithfulness ratio drop, indicating that the mechanisms are complementary rather than redundant. Besides, replacing the self-judging subagent with a far stronger external judge model (i.e., Qwen3-VL-235B-A22B) leaves accuracy essentially unchanged yet lifts faithfulness. This decoupling shows that the self-judging already can keep the accuracy without any external dependency at inference, while the remaining faithfulness gap marks an upper bound that a stronger judge can approach.

\begin{figure}[t]
    \centering
    \subcaptionbox{Accuracy\label{fig:lamb1}}[0.32\textwidth]{
        \includegraphics[width=\linewidth]{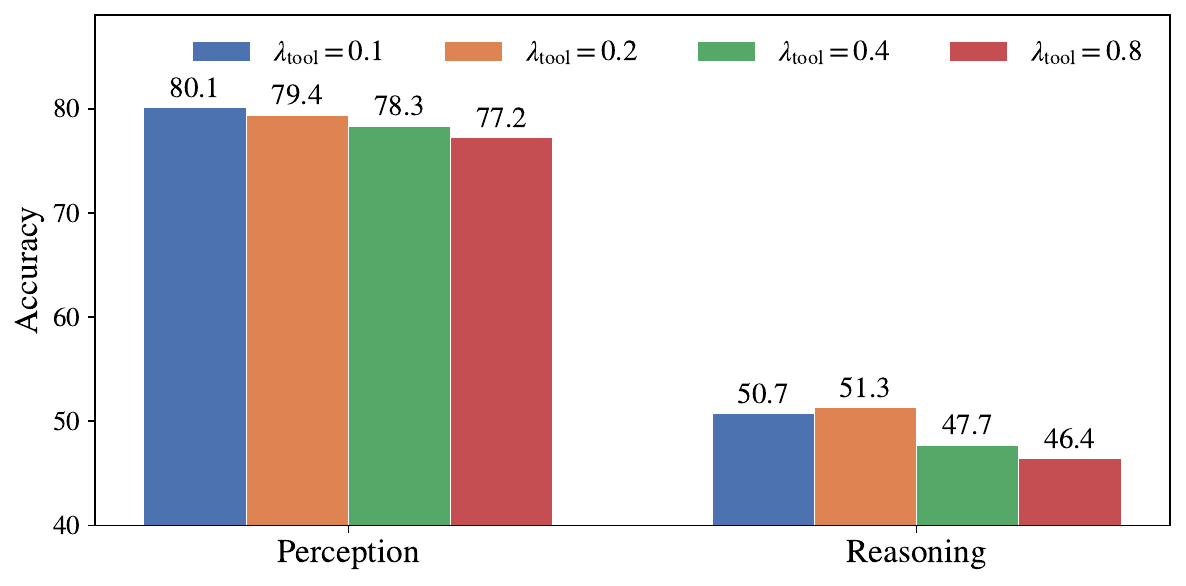}}
    \hfill
    \subcaptionbox{Tool Faithfulness Ratio\label{fig:lamb2}}[0.32\textwidth]{
        \includegraphics[width=\linewidth]{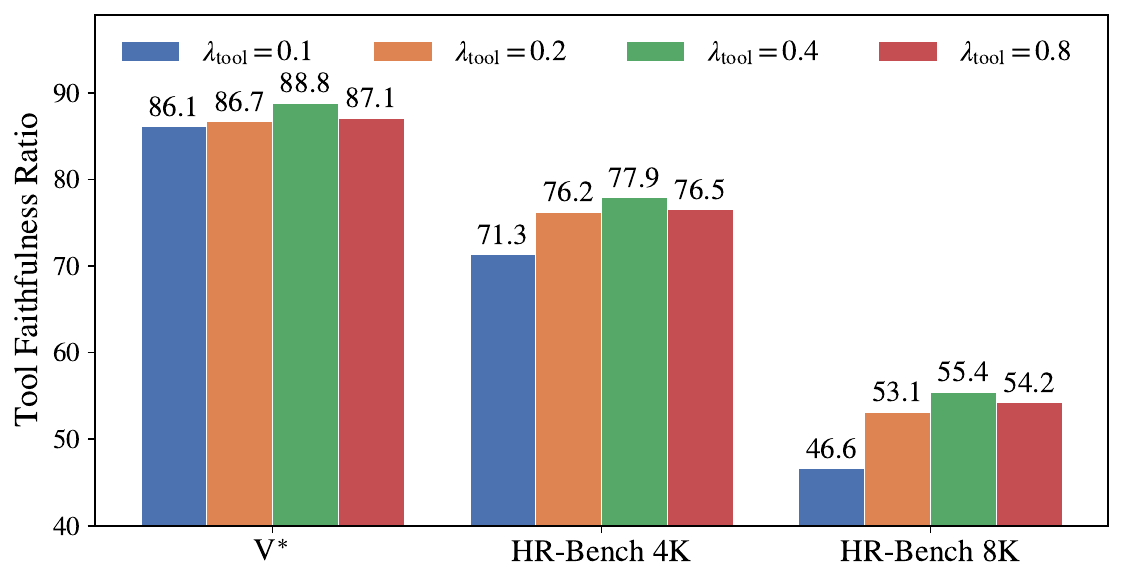}}
    \hfill
    \subcaptionbox{Tool Call Number\label{fig:lamb3}}[0.32\textwidth]{
        \includegraphics[width=\linewidth]{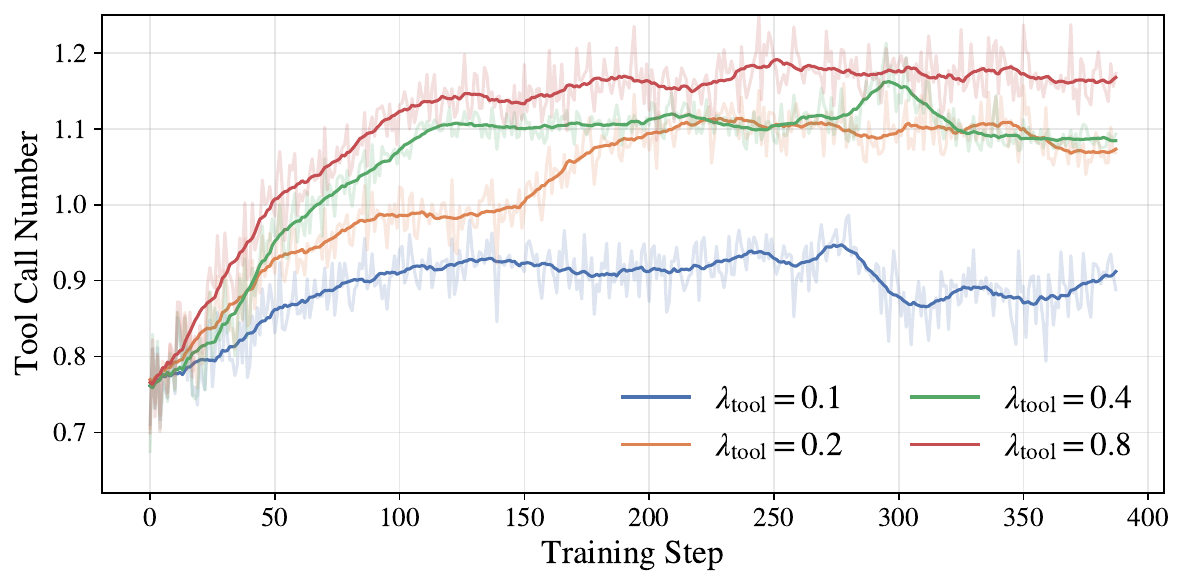}}
    \caption{Accuracy, tool faithfulness ratio and tool call number with different coefficients $\lambda_{\mathrm{tool}}$.}
    \label{lambda}
    \vspace{-5mm}
\end{figure}
\paragraph{Tool reward coefficient $\lambda_{\mathrm{tool}}$.}
Figure~\ref{lambda} ablates $\lambda_{\mathrm{tool}}\in\{0.1,0.2,0.4,0.8\}$ with $\lambda_{\mathrm{fmt}}=\lambda_{\mathrm{cons}}=0.2$ fixed, and reports accuracy, tool faithfulness ratio, and tool call number. The method is robust over a wide range. Accuracy reaches a high point at $\lambda_{\mathrm{tool}}=0.2$ and degrades at larger values, most sharply for reasoning, since an over-weighted tool term rivals the accuracy signal and lets the policy trade correctness for judge-pleasing calls. Tool faithfulness rises with $\lambda_{\mathrm{tool}}$ and saturates beyond $0.2$. Besides, the average tool call number stays around one across the entire ablation. Since $r_{\mathrm{tool}}$ is a helpfulness ratio rather than a per-call sum, increasing $\lambda_{\mathrm{tool}}$ does not directly incentivize higher call frequency, but instead encourages each call to be more useful and better targeted. We adopt $\lambda_{\mathrm{tool}}=0.2$, the point where faithfulness is already near saturation and accuracy has not yet degraded.

\section{Conclusion}
In this work, we aim to alleviate the unfaithful tool-use problem of agentic VLMs, which we attribute to two coupled shortages: an undifferentiated tool reward that credits call presence over usefulness, and a usefulness-agnostic tool observation that never signals whether the retrieved evidence is relevant or not. To tackle both, we propose \emph{\textbf{FaithEyes}}, a multi-agent self-judging framework in which the model itself serves as a subagent to judge each process image it produces. The resulting verdict is injected into the tool observation to steer subsequent reasoning, and simultaneously scales the tool reward via a helpful-tool ratio to suppress reward hacking. Training with a SFT + RL pipeline on adapted open-source data, FaithEyes attains comparable or superior accuracy across visual perception and reasoning benchmarks while markedly improving tool faithfulness and reducing inference cost. Our ablations confirm that the two mechanisms are complementary rather than redundant. Besides, grounding each tool call in verifiable visual evidence makes agentic VLMs more trustworthy for safety-critical applications. Furthermore, for recursive self-improvement of agentic VLMs, FaithEyes provides an initial validation that self-judging can effectively gate tool use.

\bibliography{colm2026_conference}
\bibliographystyle{colm2026_conference}

\newpage
\appendix
\section{More training details}

\begin{figure}[t]
    \centering
    \subcaptionbox{Data source\label{fig:data1}}[0.48\textwidth]{
        \includegraphics[width=\linewidth]{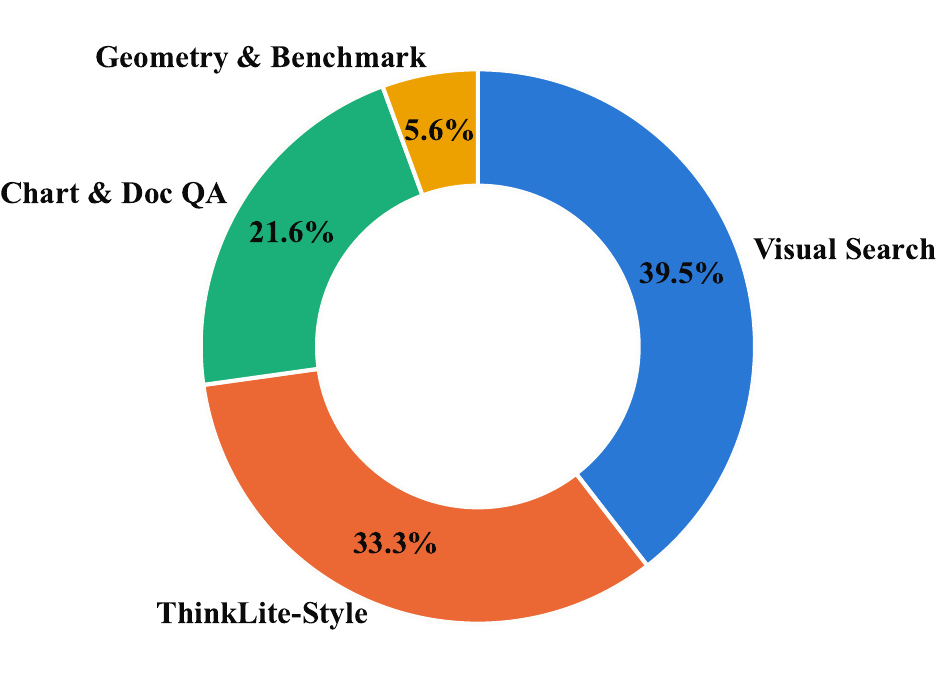}}
    \hfill
    \subcaptionbox{Data resolution\label{fig:data2}}[0.48\textwidth]{
        \includegraphics[width=\linewidth]{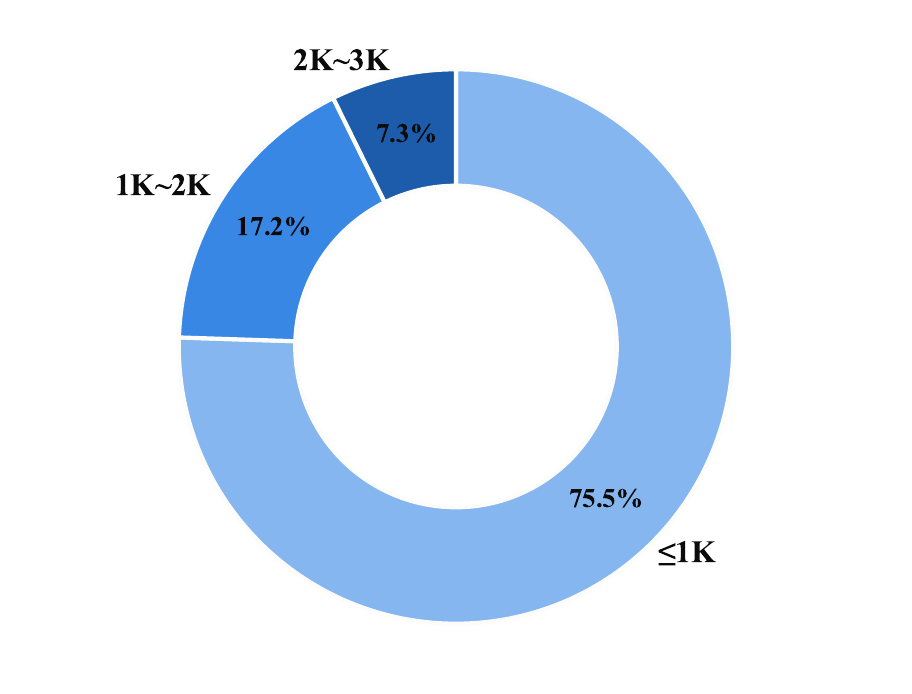}}
    \caption{RL data source and resolution distribution. Here 1K corresponds to $1008\times1008$ resolution which is 1296 visual tokens in Qwen2.5-VL, and 2K corresponds to 5184 tokens and so on.}
    \label{data_info}
    \vspace{-2mm}
\end{figure}

\paragraph{Supervised fine-tuning.}
We train all 457K SFT data for 3 epochs for Qwen2.5-VL-7B-Instruct and 5 epochs for Qwen3-VL-8B-Instruct with a batch size of 128 using AdamW~\citep{loshchilov2017decoupled} optimizer under a cosine decay schedule with a $0.05$ warmup ratio. Two masking rules are essential for stable cold-start. First, all tool observations (i.e., the returned process images together with the sandbox text output and the subagent judgement) are masked out from the loss, so that the model learns to produce code and answers rather than to predict environment feedback. Second, for two-call problem-solving trajectories, we compute the loss only on the last-call response and mask the preceding call, which prevents the model from imitating the ``deliberately-wrong-then-correct'' pattern. The main agent and subagent trajectories share the same VLM and are trained jointly in a single stage, differing only in their system and user prompts (Prompt~\ref{tmpl:main_prompt} vs.\ Prompt~\ref{tmpl:sub_prompt}). Note that, when we construct judging rationale for SFT data, we inject some hints to the user prompt of Prompt~\ref{tmpl:sub_prompt} to ensure a match with the ground truth helpful/unhelpful label. For helpful images, it is 
\begin{tcolorbox}[colback=gray!5,colframe=gray!60,boxrule=0.5pt,arc=1pt,left=1pt,right=1pt,top=1pt,bottom=1pt]
Hint: This test case should be judged as **is$\_$helpful=true**. You need to provide a suitable reason, and the reason **must not** mention judging based on the hint.
\end{tcolorbox}
For unhelpful images, it is 
\begin{tcolorbox}[colback=gray!5,colframe=gray!60,boxrule=0.5pt,arc=1pt,left=1pt,right=1pt,top=1pt,bottom=1pt]
Hint: This test case should be judged as **is$\_$helpful=false**. You need to provide a suitable reason, and the reason **must not** mention judging based on the hint. For the image containing target objects, you **must** prioritize explaining the reasons from the perspective of imprecise cropping, such as not focusing enough or the inclusion of other irrelevant targets.
\end{tcolorbox}

\paragraph{Reinforcement learning.}
After the accuracy-based filtering described in \S\ref{sec:exp:setup}, the Thyme-55K~\citep{zhang2025thyme} and DeepEyes-47K~\citep{zheng2025deepeyes} data are reduced to 50K samples. In fact, only a subset of the filtered DeepEyes data is used to keep the volume comparable to the baselines. As shown in Figure~\ref{fig:data1}, these RL samples can be grouped into four categories by question type: (i) \textit{Visual Search} data focuses on locating and grounding visual evidence, with the high-resolution subset stressing detail-intensive perception; (ii) \textit{ThinkLite-Style}~\citep{wang2026sota} data contains lightweight chain-of-thought reasoning problems; (iii) \textit{Chart \& Doc QA} data addresses chart, figure, and table reading; (iv) \textit{Geometry \& Benchmark} covers geometry and other specialized benchmarks. As shown in Figure~\ref{fig:data2}, these images span a range of resolutions. The majority of our RL data ($75.5\%$) is at or below $1$K resolution, $17.2\%$ lies between $1$K and $2$K, and the remaining $7.3\%$ between $2$K and $3$K, keeping most samples computationally affordable while a non-trivial fraction exercises high-resolution tool-calling behavior. We train the model using a fixed learning rate of $1e^{-6}$ and batch size of 128 with 12 rollouts per prompt, where each trajectory is rolled out with a sampling temperature of $1.0$ and top-$p=1.0$, a per-response cap of 20,480 tokens, and at most $5$ tool calls per trajectory to bound the interaction length. All model-generated code is run in an isolated Python sandbox with read-only access to the input image. The sandbox (i) statically scans for and blocks dangerous file operations and enforces a strict wall-clock timeout per call, (ii) normalizes working directories, auto-formats code, clamps out-of-range crop coordinates, and pre-imports common vision libraries to reduce the coding burden of a 7/8B model, and (iii) loads newly generated image files as observations. All training is conducted on 32 NVIDIA H800 GPUs, with the SFT stage taking $\sim$247 GPU hours and the RL stage taking $\sim$1361 GPU hours.

\begin{table*}[t]
\centering
\caption{Attention adjustment induced by the judgement signal in FaithEyes. Left: top-effective attention score ($\times1e^4$) from answer tokens to different context region. Right: number of samples whose attention increase ($\uparrow$) vs. decrease ($\downarrow$) on helpful and unhelpful process images after the judgement signal is injected to tool observation.}
\scalebox{0.62}{
\begin{tabular}{l|cccccc|cc}
\toprule
Models & Init img & Question & Helpful Img & Unhelpful Img & Judgement & Reasoning & Helpful $\uparrow$/$\downarrow$ & Unhelpful $\uparrow$/$\downarrow$ \\
\midrule
\multicolumn{9}{c}{\textit{V$^{*}$}} \\
\midrule
FaithEyes (w/o judgement) & 138.4 & 841.4 & 54.1 & 10.1 & 0.0 & 864.1 & \multirow{2}{*}{96 / 34} & \multirow{2}{*}{6 / 15} \\
FaithEyes                  & 132.3 & 781.8 & 58.4$\uparrow$ & 7.5$\downarrow$ & 133.7 & 911.4 &  & \\
\midrule
\multicolumn{9}{c}{\textit{HR-Bench 4K}} \\
\midrule
FaithEyes (w/o judgement) & 142.9 & 894.4 & 105.9 & 12.1 & 0.0 & 694.4 & \multirow{2}{*}{397 / 116} & \multirow{2}{*}{58 / 112} \\
FaithEyes                  & 140.2 & 844.9 & 117.2$\uparrow$ & 8.8$\downarrow$ & 145.4 & 729.5 &  &  \\
\midrule
\multicolumn{9}{c}{\textit{HR-Bench 8K}} \\
\midrule
FaithEyes (w/o judgement) & 144.7 & 862.1 & 72.2 & 35.6 & 0.0 & 660.1 & \multirow{2}{*}{227 / 75} & \multirow{2}{*}{92 / 184} \\
FaithEyes                  & 141.4 & 817.3 & 80.4$\uparrow$ & 30.6$\downarrow$ & 145.7 & 670.0 &  &  \\
\bottomrule
\end{tabular}}
\label{tab:attn_adjustment}
\end{table*}
\section{Attention adjustment}
To probe how the judgement influences the attention to process images, we replay the correct-answer trajectories of FaithEyes on V$^{*}$, HR-Bench 4K and HR-Bench 8K under two conditions: \emph{w/o judgement} (tool observation carries only the images) and \emph{w/ judgement} (the subagent verdict is appended). Here we use the keep-unhelpful trajectories. We measure how strongly the answer tokens attend to six context regions, i.e., initial image, question, helpful/unhelpful process images, judgement text and reasoning text, via attention rollout~\citep{abnar2020quantifying}. Since rolling out over all layers drives the aggregated map toward a near-uniform distribution, we accumulate only the last four layers where answer-relevant routing is most pronounced. For each region, we calculate the \emph{top-effective} attention that sums the mass on the top-$k{=}\mathrm{round}(\exp(H))$ most attended tokens per region and $H$ is the internal attention entropy of each region. This avoids both length bias and tail dilution on a common absolute scale. As shown in Table~\ref{tab:attn_adjustment}, injecting the judgement consistently raises attention to helpful images and lowers it on unhelpful ones across all three benchmarks. The per-sample paired counts also corroborate this where increasing samples outnumber decreasing ones for helpful images and the opposite phenomenon holds for unhelpful images. In fact, even without the judgement, helpful images already draw far more attention than unhelpful ones, indicating an intrinsic relevance signal that the verdict amplifies. Note that this adjustment should be read as the judgement steering attention toward the more relevant evidence, not as evidence that the answer strongly depends on the process images. In absolute terms, all images still receive far less attention than the textual regions due to the text-centric bias of VLMs, and attention is only a correlational proxy for reliance. Finally, the judgement text itself attracts non-trivial attention, indicating that the model actively incorporates its content into the reasoning chain.

\begin{figure}[t]
    \centering
    \subcaptionbox{Accuracy Reward\label{fig:sub1}}[0.32\textwidth]{
        \includegraphics[width=\linewidth]{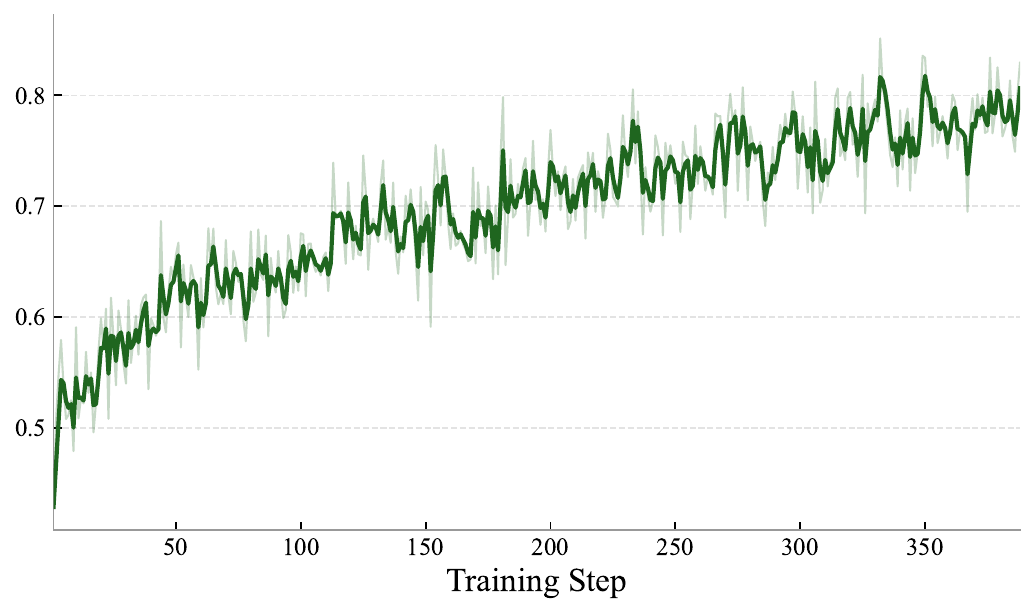}}
    \hfill
    \subcaptionbox{Format Reward\label{fig:sub2}}[0.32\textwidth]{
        \includegraphics[width=\linewidth]{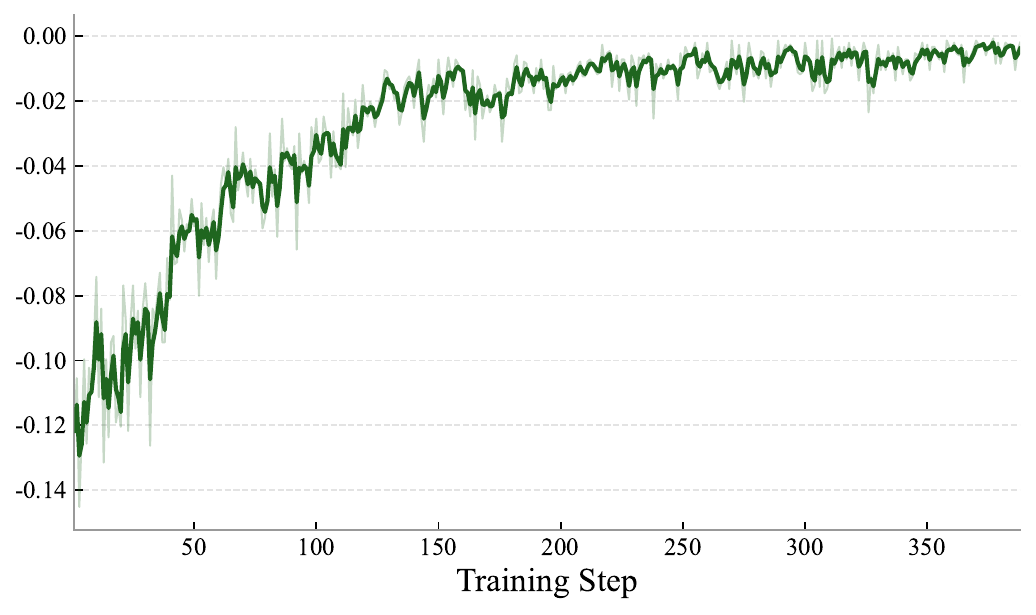}}
    \hfill
    \subcaptionbox{Consistency Reward\label{fig:sub3}}[0.32\textwidth]{
        \includegraphics[width=\linewidth]{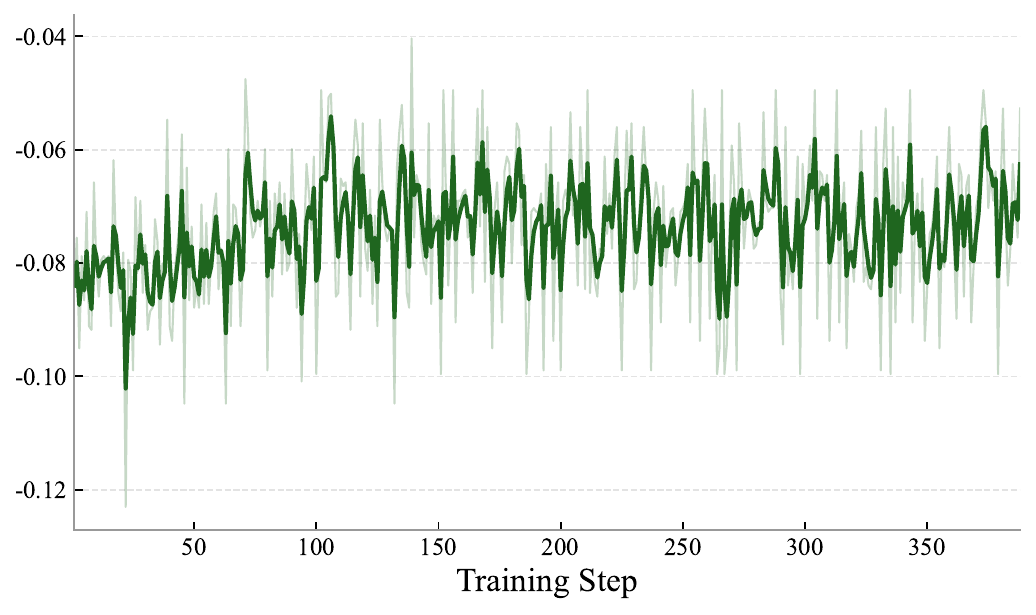}}
    \\
    \subcaptionbox{Tool Reward\label{fig:sub4}}[0.32\textwidth]{
        \includegraphics[width=\linewidth]{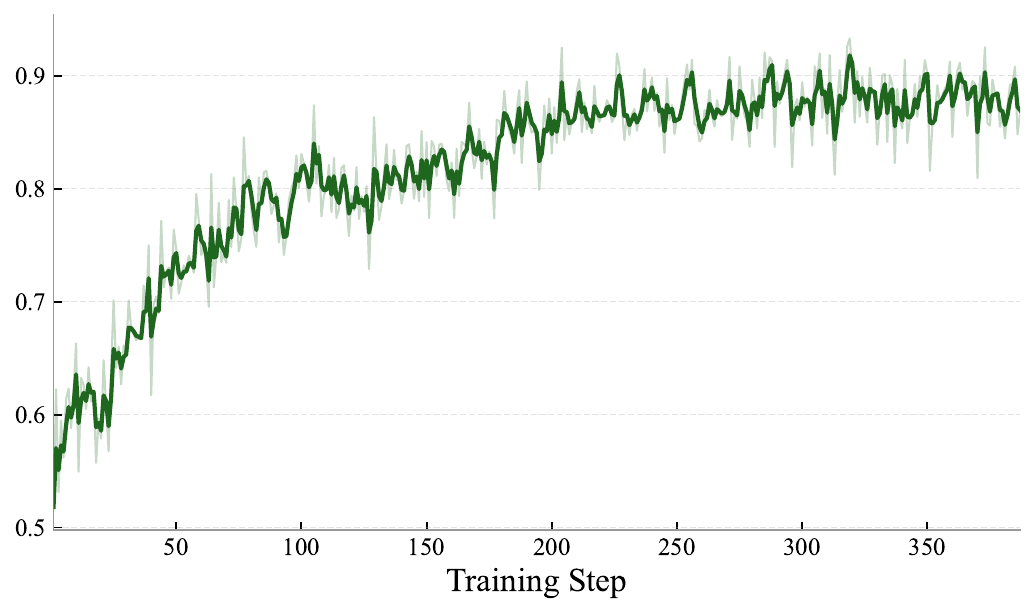}}
    \hfill
    \subcaptionbox{Tool Call Number\label{fig:sub5}}[0.32\textwidth]{
        \includegraphics[width=\linewidth]{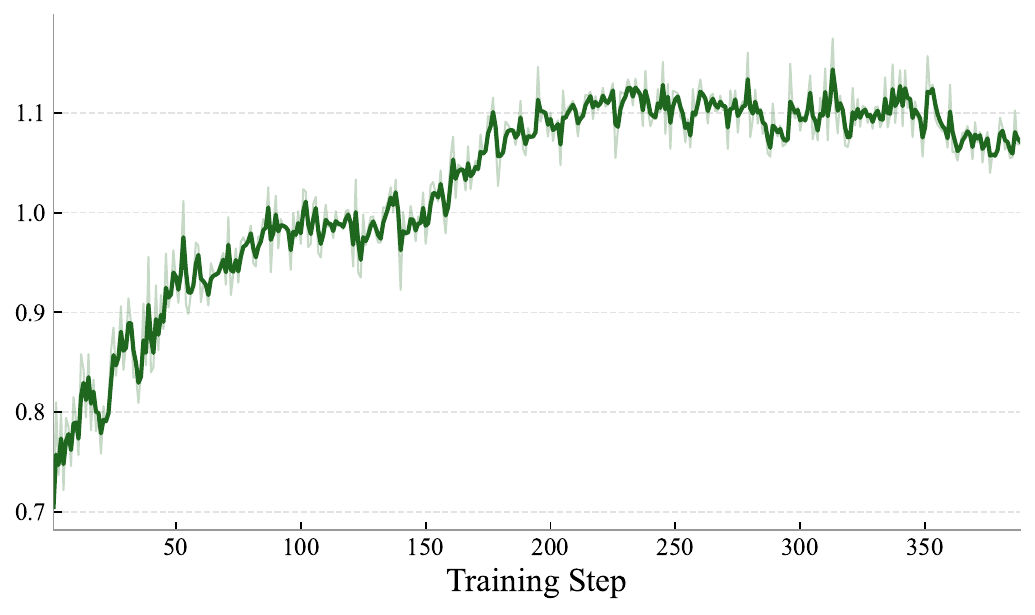}}
    \hfill
    \subcaptionbox{Response Length\label{fig:sub6}}[0.32\textwidth]{
        \includegraphics[width=\linewidth]{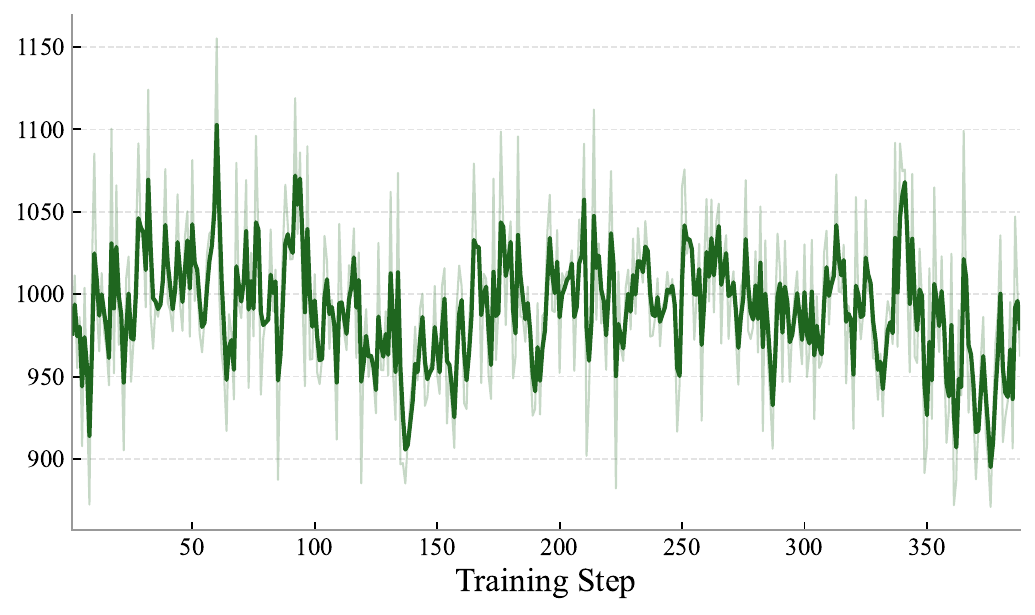}}
    \caption{Training dynamics of accuracy reward, format reward, consistency reward, tool reward (i.e., helpful-tool ratio), tool call number and response length. All scores are averaged over the rollouts within the batch.}
    \label{rl_dynamics}
\end{figure}

\section{Training dynamics}
We track the reward components and behavioral statistics throughout the RL stage to understand how FaithEyes evolves, as reported in Figure~\ref{rl_dynamics}. We plot six curves against the training step: the accuracy, format, and consistency rewards, the tool reward (i.e., the helpful-tool ratio $r_{\mathrm{tool}}$), the average number of tool calls per trajectory, and the average response length. All four reward terms improve steadily and then plateau, indicating a stable learning process without collapse. The accuracy reward rises as the policy answers more questions correctly, while the format reward climbs fastest and quickly saturates near its maximum, showing that the prescribed \texttt{\textless think\textgreater}/\texttt{\textless code\textgreater}/\texttt{\textless answer\textgreater} structure is reliably produced and parsed early in training. The consistency reward increases in tandem, meaning the final answer is increasingly entailed by the preceding reasoning rather than appended as an unsupported guess. Most relevant to our objective, the tool reward (i.e., the helpful-tool ratio) grows steadily and stabilizes at a high level, so an increasing fraction of the invoked tools are both executable and judged helpful by the subagent. Since $r_{\mathrm{tool}}$ is scaled by this ratio, its rise directly reflects that the policy learns to output faithful tool calls rather than decorative ones. The average number of tool calls gradually increases and converges to roughly one call per trajectory. Rather than over-invoking tools to farm a flat call bonus as encouraged by usage-centric rewards or collapsing into tool avoidance, the model settles on issuing about one focused and genuinely helpful tool call so that the question benefits from additional visual evidence. This is consistent with the reward design that credits usefulness rather than call frequency. Meanwhile, the average response length stays essentially flat over training, indicating that the accuracy and faithfulness gains do not stem from length hacking (e.g., padding the reasoning with verbose or repetitive text), but from more faithful on-target tool calls. Together, these dynamics show FaithEyes improves answer correctness and tool faithfulness simultaneously while keeping the reasoning concise and the tool budget small.

\begin{figure}[t]
    \centering
    \subcaptionbox{Tool Failure Ratio\label{fig:fail1}}[0.48\textwidth]{
        \includegraphics[width=\linewidth]{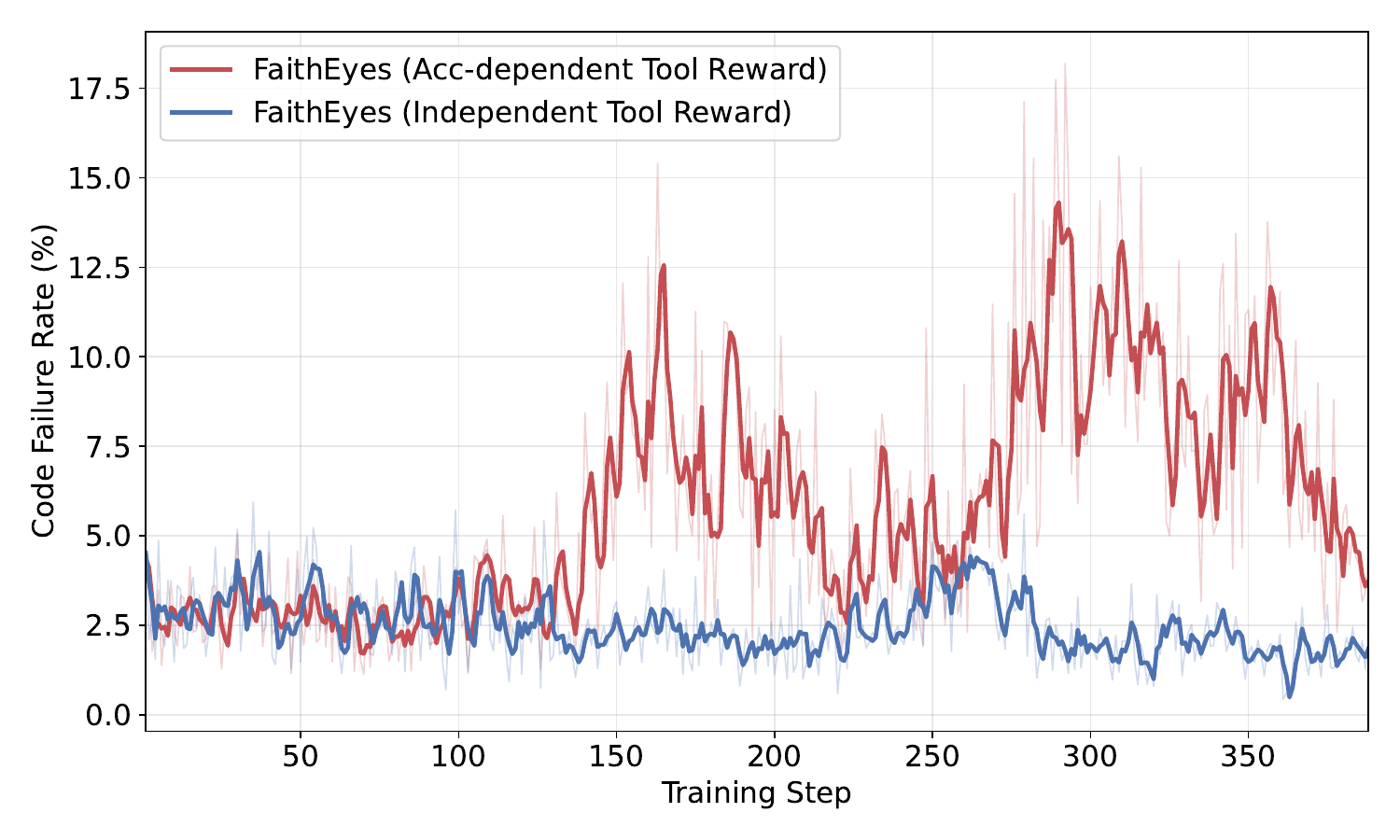}}
    \hfill
    \subcaptionbox{Tool Call Number\label{fig:fail2}}[0.48\textwidth]{
        \includegraphics[width=\linewidth]{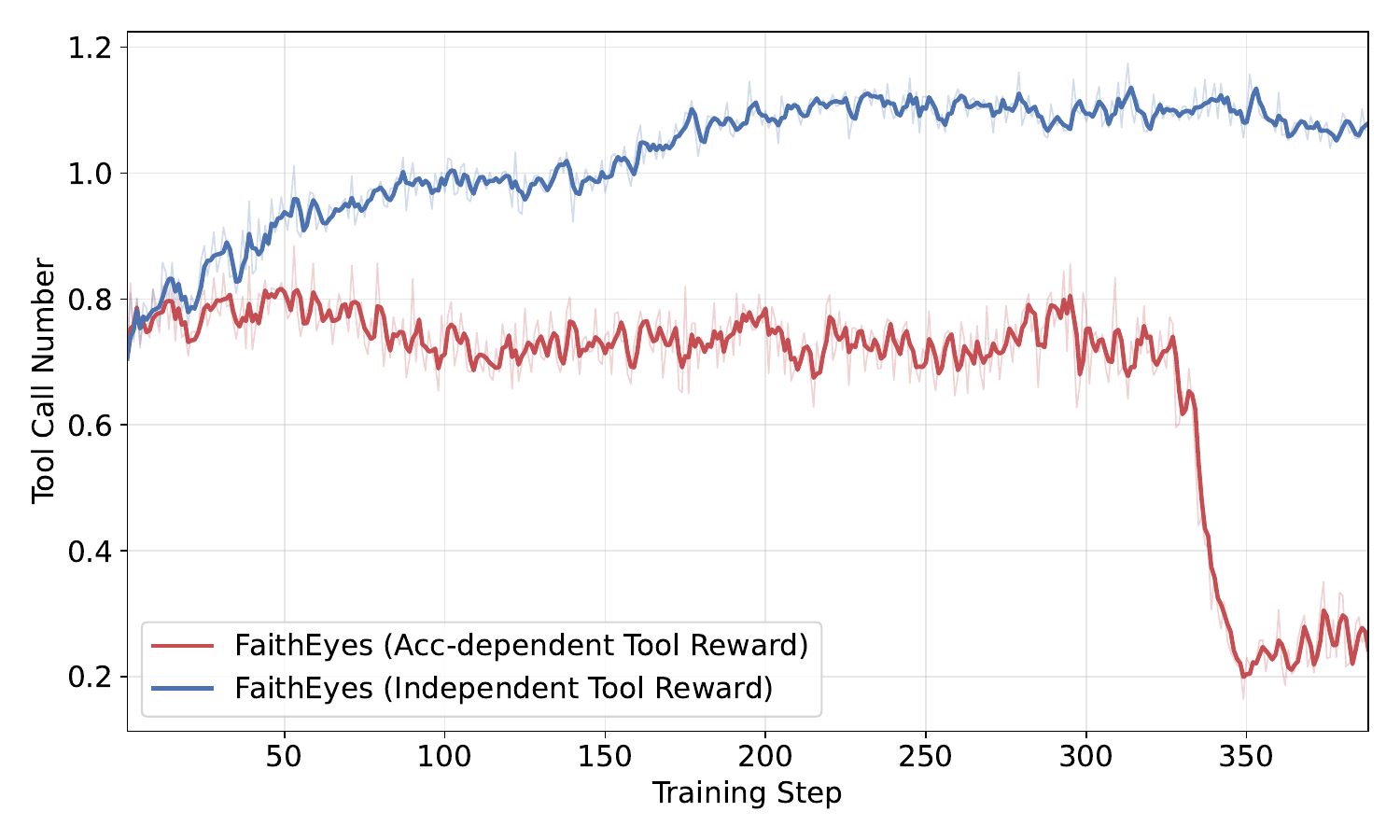}}
    \caption{Tool failure ratio and tool call number for acc-dependent and independent tool rewards.}
    \label{code_failure}
    \vspace{-2mm}
\end{figure}
\section{Why independent tool reward}
A natural design choice, adopted by several prior works~\citep{zheng2025deepeyes,yang2025thinking}, is to grant the tool bonus only when the final answer is correct. As stated in \S\ref{sec:faithEyes}, FaithEyes deliberately decouples the tool reward from the accuracy reward and computes $r_{\mathrm{tool}}$ purely from the helpful-tool ratio, independent of whether the trajectory ends with a correct answer. Here we provide the empirical evidence behind this choice. We compare two variants during RL: \emph{FaithEyes (Acc-dependent Tool Reward)}, in which the tool reward is granted only on correct trajectories, and \emph{FaithEyes (Independent Tool Reward)}, which scores tool calls on their own. Figure~\ref{code_failure} tracks their tool execution failure ratio and average number of tool calls throughout training. As shown in Figure~\ref{fig:fail1}, the acc-dependent variant exhibits two pronounced spikes in the tool execution failure ratio, peaking at roughly $18\%$, whereas the independent variant keeps the failure ratio low and stable throughout. The reason is that when the bonus is conditioned on correctness, the tool reward becomes strongly entangled with the accuracy signal. On a hard question that the model can not answer correctly, no positive tool reward is available regardless of how well the tool is used, so there is no gradient pressure to keep the emitted code executable, and the policy is free to drift toward malformed or non-executable tool calls without penalty. The independent reward, by contrast, always credits an executable and helpful call and always penalizes a failed one, providing a stable, correctness-agnostic signal that continuously discourages broken code. Figure~\ref{fig:fail2} reveals an even more damaging effect. Under the acc-dependent reward, the average number of tool calls drops sharply toward zero and never recovers. The policy degenerates into answering directly without invoking any tool. Finally, these two observations confirm that decoupling the tool reward from answer correctness is essential, as it prevents both the failure-rate blow-up and the collapse of tool use.

\begin{table*}[t]
\centering
\caption{Performance on visual perception and reasoning benchmarks.}
\scalebox{0.82}{
\begin{tabular}{l|ccc|ccc}
\toprule
      & \multicolumn{3}{c|}{\textbf{Perception}} & \multicolumn{3}{c}{\textbf{Reasoning}} \\
Model &  V$^{*}$ & HR-Bench 4K & HR-Bench 8K & MathVista & MathVerse & MathVision \\
\midrule
Qwen2.5-VL-7B-Instruct & 75.0 & 68.6 & 63.6 & 67.9 & 45.5 & 21.4 \\
FaithEyes-SFT          & 79.1 & 72.5 & 66.8 & 69.8 & 46.3 & 29.6 \\
FaithEyes-RL           & 87.4 & 77.8 & 72.9 & 73.1 & 51.0 & 29.9 \\
\midrule
Qwen3-VL-8B-Instruct   & 86.4 & 78.9 & 74.6 & 77.2 & 62.1 & 53.9 \\
FaithEyes-SFT          & 87.4 & 81.0 & 75.1 & 76.7 & 62.5 & 40.8 \\
FaithEyes-RL           & 89.9 & 83.2 & 78.5 & 78.7 & 67.6 & 46.1 \\
\bottomrule
\end{tabular}}
\label{tab:sft_rl}
\end{table*}

\begin{table*}[t]
\centering
\caption{Tool faithfulness ratio of FaithEyes-RL (Qwen3-VL-8B-Instruct) on V$^{*}$, HR-Bench 4K and HR-Bench 8K under the \emph{keep-unhelpful} and \emph{drop-unhelpful} protocols.}
\scalebox{0.94}{
\begin{tabular}{l|ccc|ccc}
\toprule
 & \multicolumn{3}{c|}{\textbf{Keep-unhelpful}} & \multicolumn{3}{c}{\textbf{Drop-unhelpful}} \\
Model & V$^{*}$ & HR-Bench 4K & HR-Bench 8K & V$^{*}$ & HR-Bench 4K & HR-Bench 8K \\
\midrule
FaithEyes-8B & 91.6 & 81.6 & 70.8 & 96.1 & 94.6 & 91.2 \\
\bottomrule
\end{tabular}}
\label{tab:faith_8b}
\end{table*}

\section{Training stage ablation}
Table~\ref{tab:sft_rl} isolates the contributions of the two training stages. Starting from Qwen2.5-VL-7B-Instruct, the cold-start SFT stage yields consistent gains across all six benchmarks ($+3.2{\sim}4.1$ points on perception tasks, $+0.8{\sim}8.2$ points on reasoning tasks). However, the SFT model merely imitates demonstrations without autonomously discriminating helpful from unhelpful tool calls. Building on this initialization, the RL stage further improves perception by $+5.3{\sim}8.3$ points, with the most pronounced gains on V$^{*}$ ($+8.3$) and HR-Bench 8K ($+6.1$), where the targets are small objects in high-resolution images that typically benefit from a focused tool call on the correct region. On reasoning tasks, FaithEyes-RL improves MathVista ($+3.3$) and MathVerse ($+4.7$), while MathVision remains essentially flat ($+0.3$). The reason may be that the dense numerical and logical reasoning in MathVision leaves little room for visual tools to help. For stronger Qwen3-VL-8B-Instruct initialization, the cold-start SFT stage still yields consistent perception gains ($+0.5{\sim}2.1$ points) and slightly lifts MathVerse ($+0.4$), yet it slightly degrades MathVista ($-0.5$) and causes a pronounced drop on MathVision ($-13.1$) and the reason may be the data distribution problem. Encouragingly, the RL stage largely repairs this damage. Optimizing with GRPO lets the model re-explore its own solving strategies rather than merely imitate demonstrations, pushing MathVista beyond both the SFT model and the base model ($78.7$ vs.\ $77.2$), lifting MathVerse by $+5.5$ over the base, and recovering $+5.3$ points of the MathVision drop. Beyond accuracy, we further examine the tool faithfulness of FaithEyes-8B in Table~\ref{tab:faith_8b}, which reports the tool faithfulness ratio on V$^{*}$, HR-Bench 4K and HR-Bench 8K under both the \emph{keep-unhelpful} and \emph{drop-unhelpful} protocols. Under the stricter keep-unhelpful protocol, where no self-judged-unhelpful images are dropped, the ratio already reaches 91.6, 81.6 and 70.8 on these three benchmarks, and dropping the unhelpful images further lifts it to 96.1, 94.6 and 91.2. This confirms that the faithfulness gains of FaithEyes stem from a genuinely more faithful tool-use policy and transfer to the stronger initialization. These results confirm that the SFT stage provides a necessary cold start, while the RL stage, guided by the faithfulness-targeting tool reward, drives the model to output genuinely useful tool calls rather than decorative ones.

\clearpage
\begin{prompttemplate}[label={tmpl:main_prompt}]{Main Agent Prompt}
\textbf{\texttt{[System]}}

You are a helpful assistant.\\

Solve the following problem step by step, and optionally write Python code for image manipulation to enhance your reasoning process. The Python code will be executed by an external sandbox, and the processed image or result (wrapped in <sandbox\_output></sandbox\_output>) can be returned to aid your reasoning and help you arrive at the final answer.\\

**Reasoning \& Image Manipulation (Optional but Encouraged):**\\
    * You have the capability to write executable Python code to perform image manipulations (e.g., cropping to a Region of Interest (ROI), resizing, rotation, adjusting contrast, drawing auxiliary lines) or perform calculation for better reasoning.\\
    * The code will be executed in a secure sandbox, and its output will be provided back to you for further analysis.\\
    * All Python code snippets **must** be wrapped as follows:\\
    <code>\\
    \verb|```|python\\
    \# your code.\\
    \verb|```|\\
    </code>\\
    * At the end of the code, print the path to the processed image (processed\_path) or the result for further processing in a sandbox environment.\\

A critic subagent will help judge whether the code execution result (i.e., processed image) is helpful for answering the user's original question and give the judgement and reasons in <sandbox\_output></sandbox\_output>.\\
You **must** carefully consider its judgment.
\par\medskip
\textbf{\texttt{[User]}} 

Image: \{ \}
Question: \{ \}

\#\#\# User Image Path:** \verb|"|\{file\_path\}\verb|"|\\
\#\#\# User Image Size:** \verb|"|\{image\_x\}$\times$\{image\_y\}\verb|"|\\
\#\#\# **Output Format (strict adherence required):**\\
<think>Your detailed reasoning process should go here.</think>\\
<code>Your compliant executable code should go here.</code>\\
<answer>Your final answer to the user's question goes here.</answer>
\end{prompttemplate}

\clearpage
\begin{prompttemplate}[label={tmpl:sub_prompt}]{Subagent Prompt}
\textbf{\texttt{[System]}}

You are evaluating whether a tool-called image is helpful for answering a visual question.\\

**Context:**\\
- The image is a processed version of an initial image, captured by a visual tool using operations such as: cropping, resizing, rotation, or contrast adjustment.\\
- Your job is to judge if this image successfully presents the information needed to answer the question.\\

**Guidelines for "is\_helpful":**\\

Set to 'true' if:\\
- The object/attribute asked in the question is clearly visible in the image.\\
- You can reasonably identify the target object and its relevant attributes (color, position, shape, etc.).\\
- The image contains the key visual information needed.\\
- The tool's processing (crop, resize, rotate, contrast adjustment, etc.) helps reveal or highlight the relevant information.\\
- The question can be answered based on what is visible in the image.\\

Set to 'false' if:\\
- The target object is NOT in the image at all (e.g., tool cropped the wrong area).\\
- The image completely missed the relevant object/region.\\
- What's shown in the image is unrelated to what the question asks about.\\
- The key information needed is outside the region or obscured by the processing.\\

**CRITICAL: Your response MUST be ONLY a valid JSON object in this EXACT format:**\\
\{"is\_helpful": true or false, "reasons": "Your brief reason here"\}\\

**Examples:**\\
\{"is\_helpful": true, "reasons": "The question asks about the tray's position. The image shows a tray on the right side of the frame. Though partially visible, its position is clear enough to answer."\}\\

\{"is\_helpful": true, "reasons": "The question asks about shirt color. The image shows a person's torso with a white shirt clearly visible. The color can be determined from this image."\}\\

\{"is\_helpful": false, "reasons": "The question asks about a red blanket, but the image shows only a kitchen counter with no blanket visible. The tool cropped the wrong area."\}\\

\{"is\_helpful": false, "reasons": "The question asks about elephant's tail, but the image only shows the elephant's legs without the tail. The key information is missing from this image."\}\\
\par\medskip
\textbf{\texttt{[User]}}

[Question]: \{ \}\\
The following image is a processed version of an initial image (possibly cropped, resized, rotated, or contrast-adjusted). Does this image show the object/attribute the question is asking about? Reply with ONLY one line of JSON.\\
Image: \{ \}
\end{prompttemplate}

\clearpage
\begin{prompttemplate}[label={tmpl:jud_prompt}]{Faithfulness Judging Prompt}
\textbf{\texttt{[System]}}

You are evaluating whether a tool-processed (cropped/zoomed) image is helpful for answering a visual question.\\

You will see the **Original Image** and the **Processed Image**. Compare them: identify the question-relevant target in the original image, then check whether the processed image contains that target and stays relatively focused on it.\\

**Rule:**\\
- Reply "true" if the processed image CONTAINS the question-relevant target AND is relatively focused on it. Blurriness and low resolution are all acceptable since the target itself might be small.\\
- Reply "false" only if the target is absent (wrong area / irrelevant object) or the crop is so unfocused it barely concentrates on the target (e.g., near full-frame copy with no real zoom/crop).\\

Reply with ONLY a single token: "true" or "false". No explanation, no punctuation, no other text.\\

Examples:\\
Question: "What does the distant sign say?"\\
- true (the processed image zooms to it; blurry due to small object, but contains the target and is focused on it)\\
Question: "What color is the blanket?"\\
- false (the processed image shows only a kitchen counter; the target is absent)\\
\par\medskip
\textbf{\texttt{[User]}}

[Original Image]: \{ \}\par
[Processed Image]: \{ \}\par
[Question]: \{ \}\par

Compare the two images: does the processed image contain the question-relevant target? Blurry and low-resolution still count as helpful. Reply with ONLY "true" or "false".
\end{prompttemplate}

\clearpage
\begin{prompttemplate}[label={tmpl:rewrite_prompt_helpful}]{Rewriting Prompt for Reasoning after Helpful Images}
\textbf{\texttt{[System]}}

You are a careful editor that rewrites reasoning blocks to integrate the preceding operation's result (image or judgement) while preserving meaning. You output only the rewritten text.\\
\par\medskip
\textbf{\texttt{[User]}}

You write ONE short opening sentence that references the result of the preceding code operation. This sentence will be placed in front of an existing reasoning block by the caller---you do NOT rewrite or repeat that block.\\

\#\#\# Context\\
- System prompt (the agent's role): \{ \}\\

- Original user question (image placeholders \texttt{\textless image\textgreater} refer to figures in the trajectory): \{ \}\\

\#\#\# Preceding operation result\\
- The processed image produced by the previous code operation (provided to you below as an actual image). The agent *should* be looking at this image.\\
\texttt{\textless image\textgreater}\\

- The subagent's judgement on this operation (it ACCEPTed the image as helpful):\\
\texttt{\textless sandbox\_output\textgreater}\\
\{ \}\\
\texttt{\textless/sandbox\_output\textgreater}\\

\#\#\# The agent's original reasoning block (for tone/language reference ONLY---do NOT include it in your output):\\
\texttt{\textless thinking\textgreater}\\
\{ \}\\
\texttt{\textless/thinking\textgreater}\\

\#\#\# Task\\
Write a SINGLE natural opening sentence that reacts to the judgement above, to be prepended before the agent's own reasoning. Concretely:\\
1. The sentence should briefly connect the judgement's conclusion to the processed image (e.g., "The cropped image, as the judgement noted, now clearly shows ...").\\
2. Keep it to exactly ONE sentence (occasionally two short clauses joined by a comma/semicolon are fine), and make the wording natural and varied---avoid the same canned opener every time.\\
3. Do NOT invent facts beyond the judgement, image, and the agent's own reasoning; do NOT restate the whole reasoning, just bridge from the judgement to it.\\
4. Match the language and tone of the agent's original reasoning block (e.g., first-person analytical voice).\\

Output ONLY the single reference sentence, with no \texttt{\textless thinking\textgreater}/\texttt{\textless/thinking\textgreater} tags, no quotes, no preamble, no explanation.\\
\par\medskip
Image: \{ \}
\end{prompttemplate}

\clearpage
\begin{prompttemplate}[label={tmpl:rewrite_prompt_unhelpful}]{Rewriting Prompt for Reasoning after Unhelpful Images}
\textbf{\texttt{[System]}}

You are a careful editor that rewrites reasoning blocks to integrate the preceding operation's result (image or judgement) while preserving meaning. You output only the rewritten text.\\
\par\medskip
\textbf{\texttt{[User]}}

You write ONE short opening sentence that references the result of the preceding code operation. This sentence will be placed in front of an existing reasoning block by the caller---you do NOT rewrite or repeat that block.\\

\#\#\# Context\\
- System prompt (the agent's role): \{ \}\\

- Original user question (image placeholders \texttt{\textless image\textgreater} refer to figures in the trajectory): \{ \}\\

\#\#\# Preceding operation result\\
- The subagent's judgement that was returned right before this reasoning block (there is NO processed image this time---the operation was judged unhelpful and its image discarded---so this is what the agent *should* be reacting to):\\
\texttt{\textless sandbox\_output\textgreater}\\
\{ \}\\
\texttt{\textless/sandbox\_output\textgreater}\\

\#\#\# The agent's original reasoning block (for tone/language reference ONLY---do NOT include it in your output):\\
\texttt{\textless thinking\textgreater}\\
\{ \}\\
\texttt{\textless/thinking\textgreater}\\

\#\#\# Task\\
Write a SINGLE natural opening sentence that reacts to the judgement above, to be prepended before the agent's own reasoning. Concretely:\\
1. The sentence should briefly note the judgement's conclusion and that the previous result was unhelpful (e.g., "As the judgement noted, the previous crop did not isolate the target.").\\
2. Keep it to exactly ONE sentence (occasionally two short clauses joined by a comma/semicolon are fine), and make the wording natural and varied---avoid the same canned opener every time.\\
3. Do NOT invent facts beyond the judgement, image, and the agent's own reasoning; do NOT restate the whole reasoning, just bridge from the judgement to it.\\
4. Match the language and tone of the agent's original reasoning block (e.g., first-person analytical voice).\\

Output ONLY the single reference sentence, with no \texttt{\textless thinking\textgreater}/\texttt{\textless/thinking\textgreater} tags, no quotes, no preamble, no explanation.
\end{prompttemplate}

\end{document}